\documentclass[dvipsnames]{article} %
\usepackage{colm2024_conference}

\usepackage{booktabs}
\usepackage{enumitem}
\usepackage{wrapfig}
\usepackage{algorithm}
\usepackage{algpseudocode}
\usepackage{graphicx}
\usepackage[misc]{ifsym}
\usepackage{multicol} 
\usepackage{microtype}
\usepackage{amsmath}
\usepackage{colortbl}
\usepackage[utf8]{inputenc}
\usepackage[T1]{fontenc}
\definecolor{lightgray}{rgb}{0.9,0.9,0.9}
\usepackage{caption}
\usepackage{subcaption}
\usepackage{graphicx}
\usepackage{subcaption}
\usepackage{setspace}
\usepackage{url}
\usepackage{multirow}
\usepackage{tabularx}
\usepackage{blindtext}
\usepackage{pgfplots}
\pgfplotsset{compat=1.18} 
\usepackage{tikz}
\usetikzlibrary{er,positioning,bayesnet}
\usepackage{makecell}
\usepackage{tipa}
\usepackage{siunitx}
\usepackage{nicefrac}
\usepackage{listings}
\usepackage[raster,skins, most]{tcolorbox} %
\usepackage{xltabular}
\usepackage{adjustbox}
\usepackage{xurl}
\usepackage{rotating}
\usepackage[normalem]{ulem}

\usepackage{fontawesome}
\usepackage{titletoc}

\usepackage[table]{xcolor}

\useunder{\uline}{\ul}{}

\usepackage{amsmath,amsfonts,bm}

\def\eqref#1{equation~\ref{#1}}

\def\1{\bm{1}}

\DeclareMathAlphabet{\mathsfit}{\encodingdefault}{\sfdefault}{m}{sl}
\SetMathAlphabet{\mathsfit}{bold}{\encodingdefault}{\sfdefault}{bx}{n}

\newcommand*\justify{%
  \fontdimen2\font=0.4em
  \fontdimen3\font=0.2em
  \fontdimen4\font=0.1em
  \fontdimen7\font=0.1em
  \hyphenchar\font=`\-
}

\renewcommand{\texttt}[1]{%
  \begingroup
  \ttfamily
  \begingroup\lccode`~=`/\lowercase{\endgroup\def~}{/\discretionary{}{}{}}%
  \begingroup\lccode`~=`[\lowercase{\endgroup\def~}{[\discretionary{}{}{}}%
  \begingroup\lccode`~=`.\lowercase{\endgroup\def~}{.\discretionary{}{}{}}%
  \catcode`/=\active\catcode`[=\active\catcode`.=\active
  \justify\scantokens{#1\noexpand}%
  \endgroup
}

\usepackage{makecell}
\usetikzlibrary{tikzmark}
\makeatletter
\newcommand*\myfontsize{%
  \@setfontsize\myfontsize{7}{8}%
}
\makeatother

\definecolor{uclablue}{RGB}{159, 195, 224}

\definecolor{uclagold}{RGB}{255, 240, 180}

\definecolor{aliceblue}{RGB}{255, 238, 241}

\definecolor{cadmiumgreen}{rgb}{0.0, 0.42, 0.24}

\definecolor{myred}{rgb}{0.7, 0.3, 0.0}
\definecolor{myblue}{rgb}{0.2, 0.3, 0.6}
\definecolor{babygreen}{rgb}{0.85, 0.97, 0.85}

\definecolor{purple1}{RGB}{126, 107, 196}
\definecolor{purple2}{RGB}{199, 158, 207}
\definecolor{purple3}{RGB}{214, 200, 255}
\definecolor{purple4}{RGB}{254, 240, 255}
\definecolor{codegray}{rgb}{0.5,0.5,0.5}
\definecolor{keywordpink}{rgb}{0.85, 0.2, 0.55} 
\definecolor{builtingreen}{rgb}{0.1, 0.6, 0.1}
\definecolor{deepblue}{RGB}{48, 58, 82}
\definecolor{codekeyword}{rgb}{0.85, 0.3, 0.6}   
\definecolor{codecomment}{rgb}{0.4, 0.6, 0.6}    
\definecolor{codebuiltin}{rgb}{0.2, 0.5, 0.7}    

\definecolor{mypurple}{RGB}{102,48,227}
\definecolor{deeppurple}{RGB}{115,51,161}

\newcommand{\symboletongyi}{\raisebox{0pt}{~\includegraphics[scale=0.012]{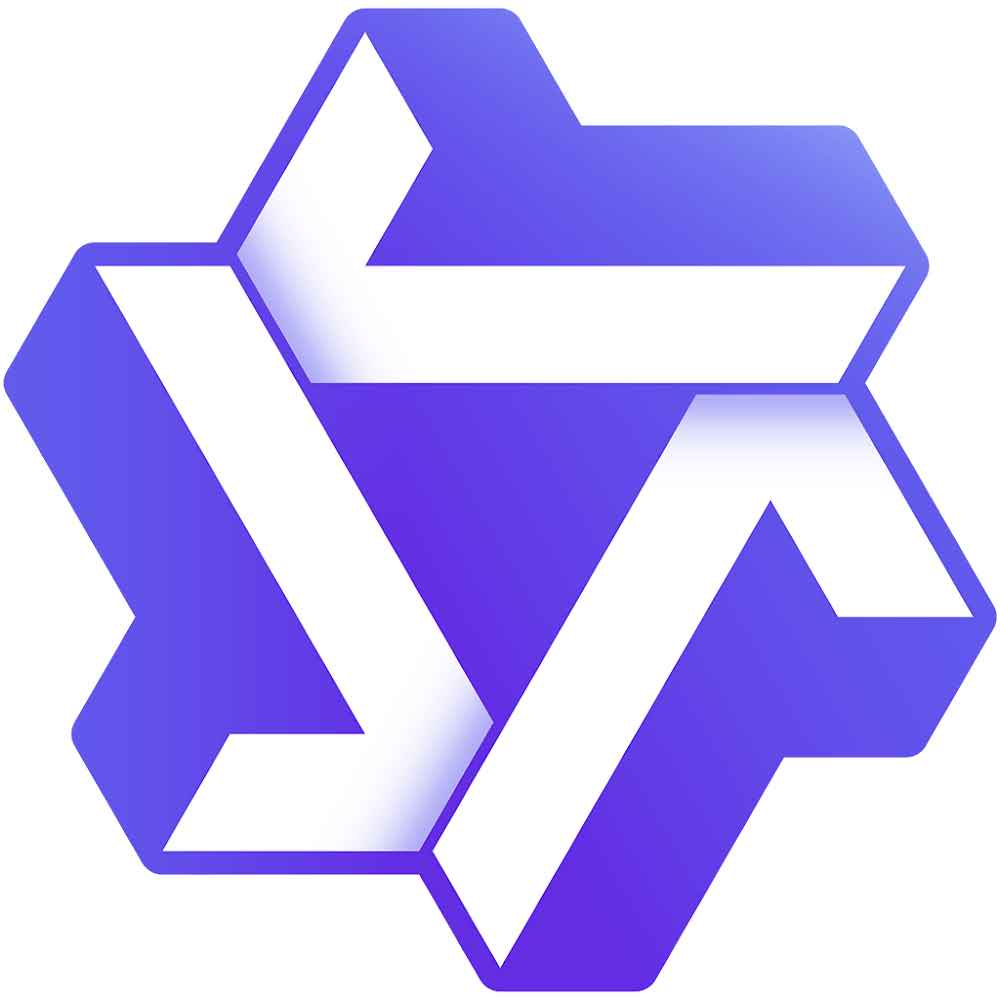}}~}

\definecolor{deepPurple}{HTML}{330066}

\definecolor{uclablue_old}{rgb}{0.15, 0.45, 0.68}
\hypersetup{
    breaklinks,
    citecolor=uclablue_old,
    colorlinks=true,
}

\newtcolorbox{mybox}[2][]
  {colback = black!5!white, colframe = black!75!black, fonttitle = \bfseries,
    colbacktitle = black!100!black, enhanced, before upper={\fontsize{8}{11}\obeyspaces\obeylines\selectfont}, fontupper=\selectfont,
    attach boxed title to top left={yshift=-2.2mm,xshift=4mm},
    title=#2,#1}

\title{%
\begin{tabular}[t]{l} 
  \parbox[t]{0.8\textwidth}{\centering 
    FIPO: Eliciting Deep Reasoning with Future-KL Influenced Policy Optimization
  }
\end{tabular}
}

\author{%
\large \symboletongyi Qwen Pilot Team, Alibaba Group \thanks{Full author list available in the \hyperref[sec:contribution]{Contributions} section.}%
  \\[1em] 
}

\begin{document}

 \maketitle

\begin{abstract}
We present \textbf{Future-KL Influenced Policy Optimization (FIPO)}, a reinforcement learning algorithm designed to overcome reasoning bottlenecks in large language models. While GRPO style training scales effectively, it typically relies on outcome-based rewards (ORM) that distribute a global advantage uniformly across every token in a trajectory. We argue that this \textbf{coarse-grained credit assignment} imposes a performance ceiling by failing to distinguish critical logical pivots from trivial tokens. FIPO addresses this by incorporating \textbf{discounted future-KL divergence} into the policy update, creating a \textbf{dense advantage formulation} that re-weights tokens based on their influence on subsequent trajectory behavior. Empirically, FIPO enables models to break through the \textbf{length stagnation} seen in standard baselines. Evaluated on Qwen2.5-32B, FIPO extends the average chain-of-thought length from roughly 4,000 to over 10,000 tokens and increases AIME 2024 Pass@1 accuracy from 50.0\% to a peak of \textbf{58.0\%} (converging at approximately 56.0\%). This outperforms both DeepSeek-R1-Zero-Math-32B ($\sim$ 47.0\%) and o1-mini ($\sim$ 56.0\%). Our results suggest that establishing dense advantage formulations is a vital path for evolving ORM-based algorithms to unlock the full reasoning potential of base models. We open-source our training system, built on the \textbf{verl} framework.
\end{abstract}

\section{Introduction}

Test-time scaling strategies such as those employed in OpenAI’s o-series \citep{jaech2024openai}, Gemini series \citep{comanici2025gemini}, and DeepSeek’s R-series \citep{guo2025deepseek} mark a fundamental shift in how large language models carry out reasoning. By allocating greater computational resources at inference time, these approaches support longer chain-of-thought and more deliberate reasoning, leading to substantial gains on demanding tasks such as competitive mathematics and coding. Much of this progress stems from large-scale reinforcement learning with verifiable rewards (RLVR) \citep{guo2025deepseek,team2025kimi,yang2025qwen3,team2025every,zeng2025glm}, which fine-tunes a model’s generation policy using feedback from task-specific verifiers, thereby eliciting and amplifying its reasoning capabilities. However, since the specific algorithms and training recipes remain largely undisclosed, it is still unclear how reinforcement learning serves as the primary catalyst to unlock potential reasoning depth, \textbf{effectively eliciting the emergence of long chain-of-thought behaviors from base models that initially exhibit no such tendencies.}

In parallel, the open-source community has devoted substantial effort to reproducing and scaling similar algorithms in more transparent settings \citep{qin2024o1,huang2024o1,liu2025understanding,hu2025open,yu2025dapo}. Among these efforts, DAPO \citep{yu2025dapo} provides a promising large-scale reproduction of GRPO-style training applied to clean base models. However, we argue that the inherent reliance on outcome-based rewards within the GRPO framework introduces a significant structural constraint. Because rewards are only binary-verifiable at the trajectory end, the standard formulation distributes a uniform advantage to every token. This results in a \textbf{completely coarse-grained credit assignment where the algorithm treats critical reasoning steps and trivial tokens with equal weight.} Specifically, we observe that reasoning trajectories produced by such baselines tend to plateau at intermediate lengths. We contend that this limitation imposes a lower performance ceiling on standard GRPO: because the uniform reward cannot highlight the specific tokens that drive correct logic, the model is unable to converge to the complex, extended reasoning paths needed for difficult tasks. While this limitation has led recent works \citep{hu2025open, yue2025vapo,fan2025truncated} to revert to the PPO framework for granular advantage estimation, we contend that such density is achievable without the complexity of a critic model.

\begin{figure}[t] 
    \centering
    \includegraphics[width=\textwidth]{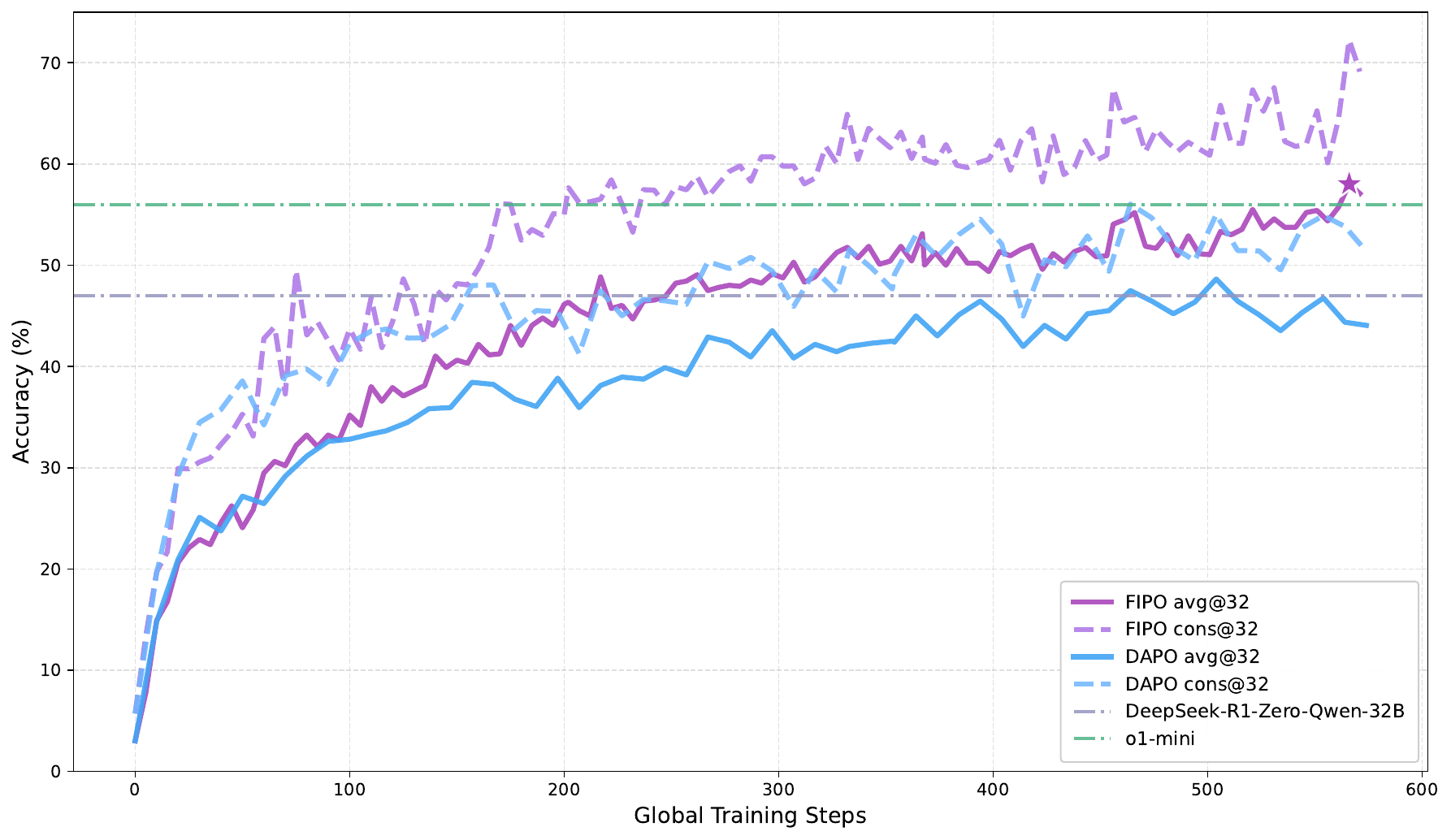}
    
    \caption{\textbf{FIPO vs. Baselines Performance Comparison on AIME2024.} FIPO demonstrates that \textbf{pure RL training alone} is sufficient to not only outperform other pure RL baselines (the reproduced DAPO and Deepseek-R1-Zero-32B), but also surpass o1-mini. This performance gain is accompanied by the generation of significantly longer responses on average.}
    \label{fig:main_comparison}
\end{figure}

We introduce \textbf{F}uture-KL \textbf{I}nfluenced \textbf{P}olicy \textbf{O}ptimization (\textbf{FIPO}). FIPO modifies the policy update by incorporating the \textbf{Future-KL divergence}, which re-weights the advantage of current tokens based on the cumulative behaviors of their subsequent trajectories. To maintain training stability, this objective is coupled with \textbf{influence weight clipping and filtering mechanism}. We evaluate this approach on \textbf{Qwen2.5-32B-Base}, a model with no prior exposure to long-CoT synthetic data, utilizing the publicly released training dataset from DAPO \citep{yu2025dapo} to ensure a strictly controlled comparison. As shown in \autoref{fig:main_comparison}, FIPO breaks the performance ceiling of standard baselines; while DAPO achieves 50.0\% (Pass@1) on AIME 2024, FIPO enables a progressive lengthening of reasoning chains, where the model steadily scales from a baseline of 4,000 tokens to a deep-reasoning regime of over 10,000 tokens. This consistent expansion pushes accuracy to a peak of 58.0\%, a result on par with recent PPO-based counterparts.\textbf{These findings demonstrate that establishing a dense advantage formulation effectively bridges the gap between GRPO efficiency and PPO performance, unlocking deep reasoning capabilities that otherwise remain untapped under uniform reward schemes.}

Our implementation is built upon the \textbf{verl} framework \citep{sheng2025hybridflow} and the DAPO codebase. By fully releasing the complete training code and configuration recipes, we aim to reveal valuable insights into large-scale reinforcement learning for LLMs that benefit the broader research community.
\section{Related Work}

\textbf{Reinforcement Learning for LLMs.} Reinforcement learning (RL) serves as a cornerstone of the post-training pipeline for large language models. While foundational efforts primarily utilized Reinforcement Learning from Human Feedback (RLHF) to align model behavior with human preferences \citep{stiennon2020learning, ouyang2022training}, recent advancements have shifted focus toward enhancing reasoning capabilities through RL. Notable examples include the OpenAI o-series \citep{jaech2024openai}, which pioneered this reasoning-centric approach, and DeepSeek-R1 \citep{guo2025deepseek}, which introduced a comprehensive RLVR \citep{lambert2024tulu} framework for developing reasoning models via the GRPO algorithm \citep{shao2024deepseekmath}. These breakthroughs have further inspired a wave of industry-leading subsequent works, such as Kimi \citep{team2025kimi}, Qwen3 \citep{yang2025qwen3}, and Gemini 2.5 \citep{comanici2025gemini}. 

\textbf{Large-scale open-source RL recipes.} Parallel to the proprietary advancements in reasoning models, the open-source community has made significant strides in democratizing large-scale RL training. These efforts aim to bridge the gap between high-level algorithmic concepts and practical, stable implementations that can scale efficiently, while providing continuous improvements to the training pipeline. Notably, GSPO \citep{zheng2025group}, BAPO \citep{xi2025bapo}, SAPO \citep{gao2025soft}, and OR1 \citep{he2025skywork} primarily develop their RL algorithms on models that have already developed long-CoT capabilities. Other works devote significant effort to incentivizing complex reasoning abilities starting from a cleaner base model, specifically \textbf{Qwen2.5-32B-Base}. Among these efforts, Open-Reasoner-Zero \citep{hu2025open}, VC-PPO\citep{yuan2025s}, VAPO \citep{yue2025vapo}, and T-PPO \citep{fan2025truncated} build their algorithms upon the PPO framework \citep{schulman2017proximal}, whereas DAPO \citep{yu2025dapo} is developed as a modification of GRPO. 

To ensure a rigorous evaluation, we adopt Qwen2.5-32B-Base as our backbone and use DAPO as our primary baseline. While Open-Reasoner-Zero reverts to PPO to avoid the sparse advantage signals in vanilla GRPO, we address this challenge by refining the GRPO framework directly. Notably, since Open-Reasoner-Zero operates without auxiliary value models, its performance ultimately falls short of DAPO. In contrast, other methods like VC-PPO, VAPO and T-PPO rely heavily on value models that are pre-trained by models already supervised fine-tuned (SFT) with Long-CoT data. We contend that this methodology introduces an external knowledge prior through the value model, creating a potential confounding factor in the evaluation. This makes it difficult to discern whether the performance gains stem from the policy optimization algorithm itself or are simply inherited from the pre-trained value model. By eschewing the need for a value model and starting from a vanilla base model, FIPO achieves performance comparable to, and in some cases superior to, these pre-trained value-model-based approaches. \textbf{This demonstrates that establishing a dense advantage formulation is a promising direction for evolving ORM-based GRPO algorithms to unlock the inherent reasoning potential of base models.}
\section{Preliminary}

In this section, we review the policy optimization frameworks central to our work: PPO and its value-network-free variants, GRPO and DAPO. Throughout this paper, let $T$ denote the total length of a trajectory and $t$ denote the index of the current step within that trajectory. In the GRPO setting, for each question prompt $q$, we sample $G$ trajectories, yielding outputs denoted by $o$.

\subsection{Proximal Policy Optimization}
\textbf{Proximal Policy Optimization (PPO)~\citep{schulman2017proximal}} introduces a clipped surrogate objective for policy optimization. By constraining policy updates to the proximity of the old policy through a clipping mechanism, PPO stabilizes training and improves sample efficiency. Specifically, PPO maximizes:
\begin{equation*}
    \mathcal{J}_{\text{PPO}}(\theta) = \mathbb{E}_{(q,o)\sim\mathcal{D}, o\sim\pi_{\theta_{\text{old}}}(\cdot|q)} \left[ \min \left( r_t(\theta)\hat{A}_t, \text{clip}(r_t(\theta), 1 - \epsilon, 1 + \epsilon)\hat{A}_t \right) \right].
\end{equation*}
Here, $r_t(\theta)= \frac{\pi_{\theta}(o_t|q,o_{<t})}{\pi_{\theta_{\text{old}}}(o_t|q,o_{<t})}$ denotes the token-level probability ratio at step $t$, $\hat{A}_t$ is the advantage estimated via a learned value function, and $\epsilon$ is the clipping coefficient. Crucially, standard PPO implementations compute the advantage $\hat{A}_t$ using Generalized Advantage Estimation (GAE) \citep{schulman2015high}. This results in distinct, token-specific advantage signals, enabling the model to perform temporal credit assignment. This stands in contrast to simplified formulations that derive advantages solely from the final outcome, effectively broadcasting a uniform signal to all tokens within a trajectory. By leveraging GAE, PPO provides dense supervision at every step, allowing it to differentiate between critical and less influential actions along the generation process.

\subsection{Group Relative Policy Optimization}

\textbf{Group Relative Policy Optimization (GRPO)~\citep{shao2024deepseekmath}} circumvents the computational burden of a value network by estimating advantages through group-based sampling. For a given query $q$ (and ground truth $a$), a set of outputs $\{o_i\}_{i=1}^G$ is sampled from the old policy $\pi_{\theta_{\text{old}}}$. The \textit{sequence-level} advantage for the $i$-th sample is standardized as:
\begin{equation}
    \hat{A}_i = \frac{R_i - \mu}{\sigma}, \quad \text{with } R_i = \mathbb{I}(\text{Verify}(o_i, a)),
\end{equation}
where $\mu$ and $\sigma$ denote the empirical mean and standard deviation, respectively, of the rewards within the sampled group. Similar to PPO, GRPO adopts a clipped objective but adds a per-token KL penalty term directly to the loss:
\begin{equation}
\begin{split}
    \mathcal{J}_{\text{GRPO}}(\theta) = \mathbb{E}_{q \sim \mathcal{D}, \{o_i\}_{i=1}^G \sim \pi_{\theta_{\text{old}}}(\cdot|q)} 
    \Bigg[ \frac{1}{G} \sum_{i=1}^G \frac{1}{|o_i|} \sum_{t=1}^{|o_i|} \bigg( & \min \Big( \rho_{i,t}(\theta)\hat{A}_{i}, \text{clip}(\rho_{i,t}(\theta), 1-\epsilon, 1+\epsilon)\hat{A}_{i} \Big) \\
    & - \beta D_{\text{KL}}(\pi_\theta || \pi_{\text{ref}}) \bigg) \Bigg]. 
\end{split}
\end{equation}
Here, $\rho_{i,t}(\theta) = \frac{\pi_\theta(o_{i,t} | q, o_{i,<t})}{\pi_{\theta_{\text{old}}}(o_{i,t} | q, o_{i,<t})}$ represents the probability ratio. By design, the computed scalar $\hat{A}_i$ is broadcast across the entire sequence; specifically, for every token $t$, the advantage is set identically as $\hat{A}_{i,t} = \hat{A}_i$. Unlike PPO, where Generalized Advantage Estimation (GAE) provides a distinct signal for each token, GRPO assigns uniform credit to every step in the trajectory, regardless of its individual contribution to the final outcome.

\subsection{Decoupled Clip and Dynamic Sampling Policy Optimization}

\textbf{Decoupled Clip and Dynamic Sampling Policy Optimization (DAPO)~\citep{yu2025dapo}} extends the GRPO framework by eliminating the explicit KL penalty. Instead, it employs asymmetric clipping within the interval $(1 - \epsilon_{\text{low}}, 1 + \epsilon_{\text{high}})$ to amplify updates for advantageous actions, effectively mitigating the entropy collapse commonly observed with GRPO. Furthermore, DAPO implements a token-level policy gradient loss to sustain healthy optimization dynamics in the context of long Chain-of-Thought RL training. Furthermore, DAPO enforces a dynamic sampling mechanism that guarantees a mix of positive and negative samples within each group $\{o_i\}_{i=1}^G$. This mechanism ensures effective updates with non-trivial gradients during optimization. We adopt DAPO as the primary baseline for this work.

\subsection{Findings on Directions of Policy Update and Fine-grained Token Analysis}

In our previous work, \citet{Meng2025sparse} provides a systematic analysis on how RL rewrites the base model. We found that in over 98\% of generation steps, the output distribution is identical. RL only intervenes at \textbf{highly sparse, critical} tokens to keep the model on track. Additionally, \citet{huang2025on} argue that Standard metrics (like KL divergence) fail to locate these sparse changes. By tracking the signed log-probability difference, we can precisely map the \textbf{``direction" of optimization}, and even boost inference accuracy just by amplifying these key tokens, with zero extra training. These insights lead to a clear conclusion: not all tokens contribute equally to the reasoning process. However, while the instantaneous log-probability difference indicates the direction of optimization, it serves merely as a primitive, localized signal. The key to eliciting more effective reasoning then lies in discovering how to leverage this raw $\Delta \log p$ to formulate a much more accurate measurement of a token's true downstream impact, thereby enabling us to automatically locate and reinforce these critical junctions during RL training.
\section{FIPO}

In this section, we introduce the core framework of FutureKL-Induced Policy Optimization (FIPO). We begin by discussing the probability shift, the fundamental building block of our objective. Next, we detail the formulation of Future-KL. Finally, we illustrate how our method implements a ``soft decay window'' strategy by focusing on the local ``future context''. This mechanism naturally prioritizes proximal signals over distant ones, limiting the effective horizon to the most relevant subsequent tokens.

\subsection{Probability Shift: $\Delta \log p$}

Our method is grounded in our recent investigations into the dynamics of Large Language Models (LLMs) during reinforcement learning. 
Specifically, our previous work on RLVR updates \citep{huang2025on} demonstrates that the magnitude and direction of the probability shift, $\Delta \log p$, serve as robust indicators of improved reasoning. 
Building upon this, our fine-grained analysis of distributional shifts \citep{Meng2025sparse} further reveals that this generation process is often driven by a few ``sparse but critical'' tokens that disproportionately influence the subsequent chain of thought. Inspired by these insights, we identify the token-level probability shift as the atomic unit for our credit assignment mechanism. Formally, we define the probability shift at time step $t$ as the log-space difference between the current policy and the old policy:
\begin{equation}
    \Delta \log p_t = \log \pi_{\theta}(o_t \mid q, o_{<t}) - \log \pi_{\theta_{\text{old}}}(o_t \mid q, o_{<t}).
\end{equation}
This term serves as a differential signal capturing the instantaneous policy drift:
\begin{itemize}
    \item \textbf{Positive Shift ($\Delta \log p_t > 0$):} Indicates that the current policy has increased the likelihood of token $o_t$ relative to the old policy. This typically suggests that the training objective is reinforcing this specific reasoning step.
    \item \textbf{Negative Shift ($\Delta \log p_t < 0$):} Implies that the policy is suppressing the generation of $o_t$, signaling that the updated model is actively down-weighting this specific token relative to the reference policy.
\end{itemize}

Unlike traditional KL penalties, which treat this drift primarily as a regularization cost to be minimized, we interpret $\Delta \log p_t$ as a directional signal of behavioral adjustment, thereby explicitly coupling the optimization objective to the generative dynamics. However, relying solely on this instantaneous shift is insufficient, as it fails to capture the long-term consequences of a decision. This limitation motivates our proposed \textbf{Future-KL} mechanism, which re-weights the current token by aggregating the distributional shifts of its \textit{future} trajectory.

\subsection{Future-KL Estimation}
\label{sec:future_kl}

While $\Delta \log p_t$ captures the local distributional shift, reasoning is inherently a sequential process where the true significance of this token depends on the trajectory it initiates. To capture this causal influence, we define \textbf{Future-KL} as the cumulative signed probability shift from the current step $t$ to the end of the sequence $T$:
\begin{equation}
    \mathrm{FutureKL}_t = \sum_{k=t}^{T} \Delta \log p_k.
\end{equation}

This summation is mathematically equivalent to the log-likelihood ratio of the joint probability distributions for the subsequent sequence $o_{t:T}$. It can thus be interpreted as a sample-based estimate of the KL divergence restricted to the future horizon, measuring the cumulative deviation of the current policy from the reference policy for the remainder of the trajectory. We therefore term this metric \textit{Future-KL}. Functionally, $\mathrm{FutureKL}_t$ serves as a forward-looking metric that quantifies the cumulative shift in policy distribution regarding the future trajectory. A positive value ($\mathrm{FutureKL}_t > 0$) indicates that the updated policy has overall \textbf{reinforced} the entire subsequent trajectory initiated by token $o_t$, suggesting that $o_t$ acts as a stable anchor for the subsequent reasoning chain. In contrast, a negative value ($\mathrm{FutureKL}_t < 0$) implies that the policy is collectively suppressing the future tokens following $o_t$, signaling that the trajectory stemming from this point is becoming less favored during the optimization process.

\begin{figure}[ht] 
    \centering
    \includegraphics[width=\textwidth]{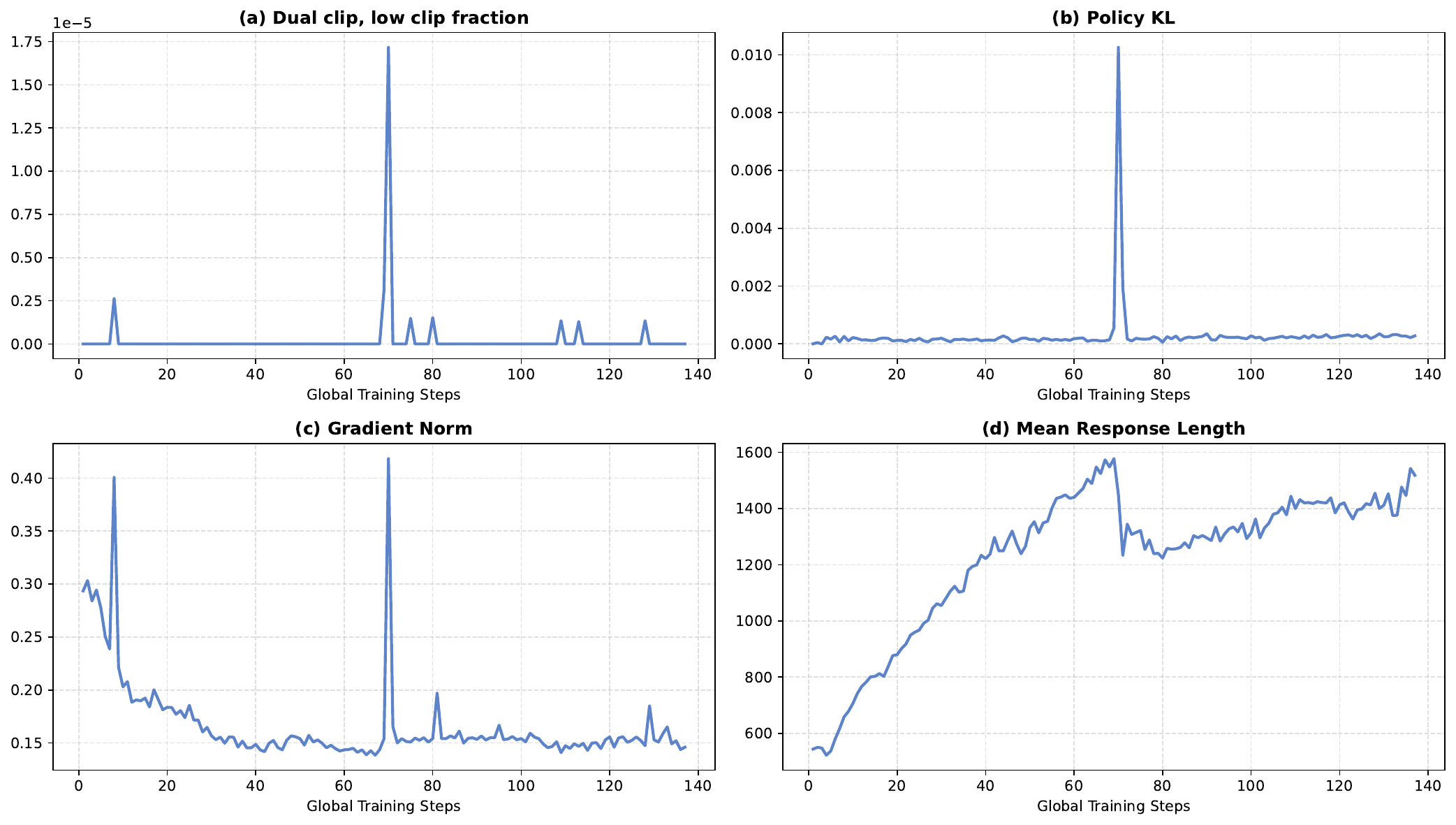}
    
    \caption{\textbf{Training instability with vanilla FutureKL.} 
    Analysis of the unstable run observed around Step 70. 
    (a) A sharp spike in the \textit{low-clip fraction} (indicating a drastic shift in the policy distribution driven by negative samples) triggers 
    (b) a sudden divergence in Policy KL. 
    (c) an immediate explosion in gradient norm and 
    These internal instabilities collectively precipitate 
    (d) a catastrophic collapse in response length, confirming that unregulated negative signals destabilize the optimization.}
    \label{fig:instability_plots}
    \vspace{-4mm}
\end{figure}

However, in practice, such formulation tends to exacerbate the variance arising from distributional shifts. Since $\mathrm{FutureKL}_t$ acts as a weighting coefficient for the advantage function (as detailed in subsequent sections), excessive deviations in future logits (e.g., due to training-inference inconsistency) can disproportionately inflate the scale. This renders the optimization overly sensitive to noisy tokens rather than the intrinsic quality of the reasoning chain. Empirically, we observe that in the absence of safety mechanisms, training runs are prone to severe instability. As shown in \autoref{fig:instability_plots}, this collapse is distinctively accompanied by a sharp spike in the ``low-clip fraction'' metric, which tracks the frequency of samples triggering the Dual-Clip threshold (a hard clip ratio on negative samples) \citep{ye2020mastering}. Such high importance ratios on negative samples signify a critical misalignment: the model assigns high probability to an action that is effectively harmful. In our experiments, this spike (at approximately Step 70) aligns with a surge in the gradient norm and Policy KL\footnote{We compute the Policy KL divergence as the batch mean of the negative log-ratio: $\text{Policy KL} = \frac{1}{B \cdot L} \sum_{i=1}^{B} \sum_{t=1}^{L} \left( \log \pi_{\text{old}}(o_{i,t}|o_{<t}) - \log \pi(o_{i,t}|o_{<t}) \right)$. It measures the KL divergence of the generated sequences between the current policy and the policy prior to the gradient update (the roll-out policy).}, indicating a substantial shift in policy distribution, alongside an immediate drop in response length. This synchronization indicates that without regulation, the accumulated negative signals from $\mathrm{FutureKL}_t$ can reach some extreme values that destabilize the training process.

Motivated by these observations, we refine the $\mathrm{FutureKL}$ computation by explicitly masking tokens that exceed the Dual-Clip threshold. Since these tokens represent 'harmful' actions whose gradients are already clipped (via the clipped policy objective), allowing their excessively high importance ratios to propagate into the recursive sum introduces severe variance. By zeroing out the future accumulation for these specific outliers, we remove the primary source of instability. The refined objective is defined as:

\begin{equation}
    \mathrm{FutureKL}_t = \sum_{k=t}^{T} M_k \cdot \Delta \log p_k, \quad
    M_k = \mathbb{I}\left( \frac{\pi_\theta(o_k|o_{<t})}{\pi_{\text{old}}(o_k|o_{<t})} \le c \right).
\end{equation}

Here, $M_k$ acts as a binary filter that evaluates to 1 only if the importance ratio remains within the Dual-Clip threshold $c$ (typically $c \geq 10$), and 0 otherwise. This ensures that tokens triggering the hard constraints are effectively excluded from the FutureKL computation, preventing gradient explosion without altering the trajectory's valid signals.

\subsubsection{Soft Decay Window}

Beyond the stability constraints, we also address the inherent uncertainty of long-horizon generation. The causal dependency between the current action $o_t$ and future tokens $o_k$ naturally diminishes as the time horizon $k-t$ increases. Immediate successors are directly conditioned on the current choice, whereas distant tokens are subject to accumulating stochasticity and become less predictable. To model this diminishing influence, we introduce a discount factor $\gamma \in (0, 1]$. Incorporating this decay into the masked objective yields the final formulation used in our experiments:

\begin{equation}
    \mathrm{FutureKL}_t = \sum_{k=t}^{T} M_k \cdot \gamma^{k - t} \cdot \Delta \log p_k.
\end{equation}

We parameterize the decay rate as $\gamma = 2^{-\frac{1}{\tau}}$, where $\tau$ is a hyperparameter controlling the effective horizon (or ``half-life'') of the future supervision. This formulation ensures that the credit assignment concentrates on the immediate reasoning chain, assigning lower weights to distant, highly uncertain tokens. Functionally, $\tau$ defines the aperture of this \textit{soft decay window}. Unlike a hard truncation that abruptly discards information beyond a fixed step, this exponential formulation creates a continuous sliding window where $\tau$ represents the distance at which the future signal's influence attenuates by half. This mechanism allows the model to prioritize local coherence within the window $\tau$, while smoothly filtering out the noise from the distant future without introducing boundary artifacts.

\subsubsection{FutureKL Re-weighted Advantage with Clipping}

Finally, we integrate the soft decay window and masking mechanisms into the policy optimization objective. We propose to modulate the standard advantage estimate $\hat{A}_t$ using a future influence weight $f_t$. The modified advantage $\tilde{A}_t$ is defined as:

\begin{equation}
    f_t = \operatorname{clip}\left( \exp(\mathrm{FutureKL}_t), \, 1 - \epsilon_{f_{low}}, \, 1 +  \epsilon_{f_{high}} \right), \quad
    \tilde{A}_t = \hat{A}_t \cdot f_t.
\end{equation}

This formulation introduces two key operations:
\begin{enumerate}
    \item \textbf{Exponential Mapping:} We transform the accumulated scalar signal from log-space to a multiplicative domain. Mathematically, the unclipped term represents a decay-weighted product of likelihood ratios, which acts as an importance weight reflecting the policy's effective preference for the generated future.
    \item \textbf{Influence Weight Clipping:} We constrain the multiplicative coefficient $f_t$ to the interval $[1-\epsilon_{f_{low}}, 1+\epsilon_{f_{high}}]$. This operation serves strictly to bound the magnitude of the advantage modulation, preventing the exponential term from introducing excessive variance into the gradient estimate. By capping the weight, we ensure that the future trajectory modulates the update signal within a controlled range, avoiding numerical instability caused by extreme accumulated log-probability shifts.
\end{enumerate}

Functionally, this modulation scales the magnitude of the policy update based on the reinforcement or suppression of the generated future. When the updated policy \textbf{reinforces} the subsequent trajectory (i.e., $\mathrm{FutureKL}_t > 0$), the weighting term $f_t > 1$ magnifies the gradient signal. Consequently, positive advantages are boosted to encourage the current token as a stable anchor, while negative advantages incur harsher penalties to strictly correct errors initiating this path. Conversely, when the policy \textbf{suppresses} the future trajectory (i.e., $\mathrm{FutureKL}_t < 0$), the term $f_t < 1$ attenuates the update. This attenuation effectively reduces the reward signal for locally harmful tokens that happen to be in a successful sequence and softens the penalty for good tokens trapped in a failing one. In practice, to ensure training stability and prevent over-penalization, we reset $f_t = 1$ for tokens associated with negative advantages ($\hat{A}_t < 0$) that exhibit excessively large importance ratios.

\subsection{Target Loss}

Adopting the token-level formulation from DAPO \citep{yu2025dapo}, we maximize the following FIPO objective:
\begin{equation}
    J_{\text{FIPO}}(\theta) = \mathbb{E}_{(q,a)\sim \mathcal{D},\,\{o_i\}\sim \pi_{\theta_{\text{old}}}} \left[ \frac{1}{\sum_{i=1}^{G}|o_i|} \sum_{i=1}^{G}\sum_{t=1}^{|o_i|} \min \left( r_{i,t} f_{i,t} \hat{A}_{i,t}, \, \operatorname{clip}\left(r_{i,t}, 1-\epsilon, 1+\epsilon\right) f_{i,t} \hat{A}_{i,t} \right) \right].
\end{equation}

Here, $G$ represents the number of sampled outputs per query, and $r_{i,t} = \frac{\pi_\theta(a_{i,t}|s_{i,t})}{\pi_{\theta_{\text{old}}}(a_{i,t}|s_{i,t})}$ denotes the importance ratio between the current and old policies. The term $\hat{A}_{i,t}$ refers to the group relative advantage, while $f_{i,t}$ serves as the Future-KL importance weight introduced previously.

\section{Experiment}

\subsection{Experiment Settings}

In this work, we adopt the training settings of DAPO \citep{yu2025dapo}, specifically focusing on mathematical reasoning tasks to ensure a strictly controlled comparison. We utilize the VeRL framework \citep{sheng2025hybridflow} for both training and baseline reproduction. We maintain optimization settings consistent with DAPO, and trained on the public-released DAPO-17K dataset. Each training batch consists of 512 prompts with 16 responses sampled per prompt, yielding a total of 8,192 training samples. In the standard DAPO configuration, updates are performed with a mini-batch size of 512 samples (32 prompts), resulting in 16 gradient updates per training iteration. \textbf{However, our empirical findings suggest that a larger mini-batch size improves training stability.} Consequently, we adopted a mini-batch size of 1,024 samples (64 prompts), resulting in 8 gradient updates per iteration. A more detailed discussion regarding the impact of this increased minibatch size is provided in Appendix Sec. \ref{app_sec: Reproduction_discussion}. For the \textbf{Future-KL computation}, we set the effective horizon of the decay rate $\tau$ to 32. Specific to the training of 32B model, the Future-KL weight is clipped within $[1, 1.2]$; \textbf{this effectively amplifies the reward for tokens associated with successful reasoning trajectories while imposing a more stringent penalty for those leading to incorrect outcomes.} Both FIPO and DAPO share a maximum response length of 20,480 tokens, with an overlong penalty applied to trajectories exceeding 16,384 tokens. Detailed hyperparameter configurations for both the baseline and FIPO are provided in Appendix Sec. \ref{sec: parameters}. 

For evaluation, we adopt AIME 2024 as our primary validation benchmark, supplemented by AIME 2025, to ensure a rigorous and comprehensive comparison with the DAPO baseline. To maintain results stability and account for variance in chain-of-thought generation, we follow the DAPO protocol by repeating the evaluation 32 times and reporting the Pass@1 (averaged over 32 samples). Inference hyperparameters are consistently set to a temperature of 1.0 and a top-$p$ of 0.7.


\subsection{Main Result}

\autoref{tab:aime_results} presents the quantitative evaluation on the AIME 2024 and AIME 2025 benchmarks. FIPO achieves a systematic improvement of roughly 6.0\% in Pass@1 (Avg@32) over the DAPO baseline across both datasets. We prioritize this metric as the most robust indicator of reasoning reliability. While we also observe gains in consistency, the improvement in coverage (Pass@32) is more modest, particularly on AIME 2025. We attribute this to the inherent challenge of expanding the absolute problem-solving scope of large models through reinforcement learning alone. Without external knowledge augmentation or tool integration, RL is primarily constrained to refining how the model navigates its existing internal knowledge. Consequently, while FIPO significantly enhances the model's ability to reliably solve problems within its latent capacity (driving up Avg@32), shifting the boundary of solvable problems (Pass@32) remains non-trivial.

\begin{table}[ht]
\centering           
\caption{\textbf{Comparison of reasoning performance on AIME benchmarks.} All results are reported as percentages (\%). We report the average Pass@1 across 32 samples (Avg@32), the majority vote (Cons@32), and the probability of at least one correct answer (Pass@32). To align with prior baseline reporting and reduce sensitivity to digit-level generation variance, final values are rounded to the nearest integer.}
\label{tab:aime_results}
\begin{tabular}{lcccccc}
\toprule
\multirow{2}{*}{\textbf{Method}} & \multicolumn{3}{c}{\textbf{AIME 2024}} & \multicolumn{3}{c}{\textbf{AIME 2025}} \\
\cmidrule(lr){2-4} \cmidrule(lr){5-7}
 & Avg@32 & Cons@32 & Pass@32 & Avg@32 & Cons@32 & Pass@32 \\
\midrule
DAPO (Baseline) & 50.0\% & 60.0\% & 80.0\% & 38.0\% & 47.0\% & 63.0\% \\
\textbf{FIPO (Ours)} & \textbf{56.0\%} & \textbf{73.0\%} & \textbf{83.0\%} & \textbf{43.0\%} & \textbf{50.0\%} & \textbf{67.0\%} \\
\bottomrule
\end{tabular}
\vspace{-3mm}
\end{table}
\section{Analysis}

Beyond the aggregate metrics, we observe several distinct phenomena that we believe underpin these performance gains. By dissecting the training dynamics and inference behaviors, we identify three critical drivers of FIPO's effectiveness: the \textbf{emergence of length-based scaling} in reasoning chains, the \textbf{distinct positive learning signal} captured by the response length weighted mean advantage formulation, and the significantly \textbf{improved stability} of the optimization process compared to standard baselines.

\subsection{The scaling of length and performance}

\begin{figure}[!ht] 
    \centering
    \includegraphics[width=\textwidth]{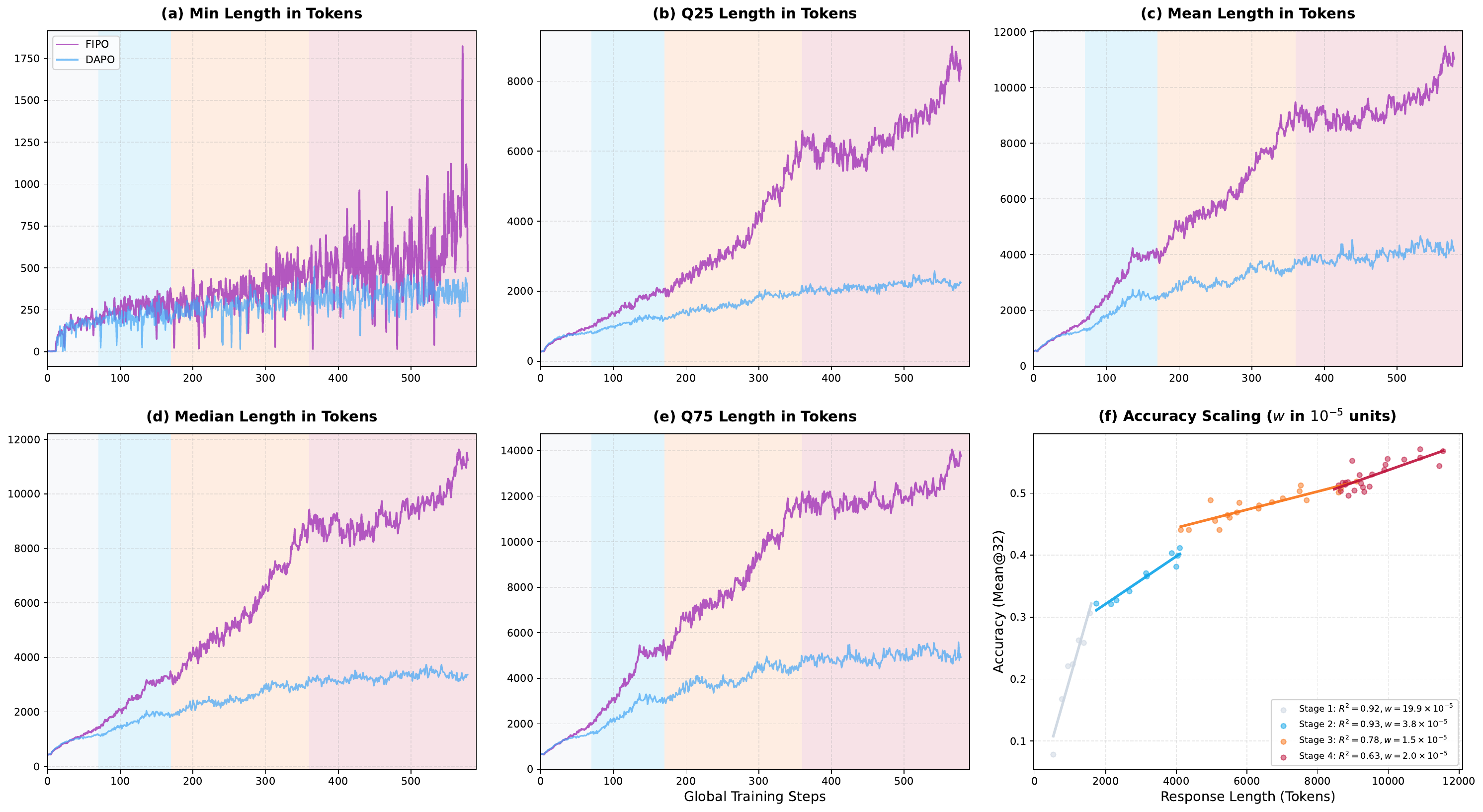}
    \caption{\textbf{Dynamics of response length and performance scaling during training.} Subplots (a)-(e) show the evolution of response length metrics (Min, Q25, Mean, Median, Q75) over global training steps. Compared to the DAPO baseline, FIPO significantly increases response length, effectively eliciting more extensive Chain-of-Thought reasoning. Subplot (f) demonstrates that this increased length correlates strongly with improved accuracy, suggesting that longer CoT is key to breaking performance barriers.}
    \label{fig:response_length}
    \vspace{-4mm}
\end{figure}

\textbf{A central observation in FIPO's training is that performance gains are deeply coupled with a continuous expansion of response length.} As training progresses, we observe a significant surge in token counts that scales alongside model accuracy. As illustrated in \autoref{fig:response_length}, the response length of DAPO gradually enters a stagnation phase after an initial increase, plateauing at an average of approximately 4,000 tokens. In contrast, FIPO exhibits remarkable scaling resilience. This scaling process unfolds through distinct evolutionary phases (visualized by the colored regions in \autoref{fig:response_length}), marking a transition from an initial rapid exploration to a sustained period of deep reasoning. Notably, although an overlong penalty is maintained to constrain redundancy, FIPO successfully guides the model to elicit extensive Chain-of-Thought (CoT) reasoning. Qualitative analysis provided in Appendix \ref{app_sec: case_study} reveals that this length expansion is driven by the gradual emergence of self-reflection behaviors; the model increasingly utilizes the expanded sequence length to re-evaluate its intermediate steps and explore multiple methodologies to verify its conclusions. Interestingly, this spontaneous emergence of systematic self-verification aligns with the inference-time scaling behaviors observed in advanced reasoning models (e.g., the OpenAI o-series and DeepSeek-R1). This suggests that FIPO effectively triggers \textbf{inference-time reasoning}, prioritizing \textbf{analytical depth} to unlock higher performance.

\textbf{Further examination of the training dynamics reveals that this surge in length is not driven by isolated outliers but represents a comprehensive distributional migration.} As shown in \autoref{fig:response_length}(a)--(e), all length-related percentiles, ranging from the Minimum and Q25 to the Median and Q75, exhibit a synchronized and stable upward shift under FIPO training. Specifically, across these training phases, the median token count climbs steadily from an initial 200 to over 10,000. Such a migration across the entire distribution demonstrates that FIPO facilitates a fundamental shift in the model's underlying problem-solving strategy: the model transitions from direct response patterns to systematic, self-verifying reasoning processes. Crucially, we find that this collective shift toward longer reasoning chains is what unlocks the performance breakthroughs observed in our experiments. As illustrated in \autoref{fig:response_length}(f), there is a strong positive correlation between model accuracy and response length across all identified stages. While the correlation slopes (denoted as $w$) vary slightly between phases, the trajectory remains consistently positive. While the DAPO baseline's performance reaches a bottleneck as its length plateaus, FIPO's ability to continuously unlock additional ``thinking space'' allows the model to navigate increasingly complex logical dependencies. This confirms that \textbf{FIPO successfully converts increased sequence length into genuine reasoning depth, enabling the model to surpass the performance ceilings of standard baselines on high-difficulty reasoning tasks.}

\subsection{The dynamics of advantage and sustained reasoning growth}
\begin{figure}[t] 
    \centering
    \includegraphics[width=\textwidth]{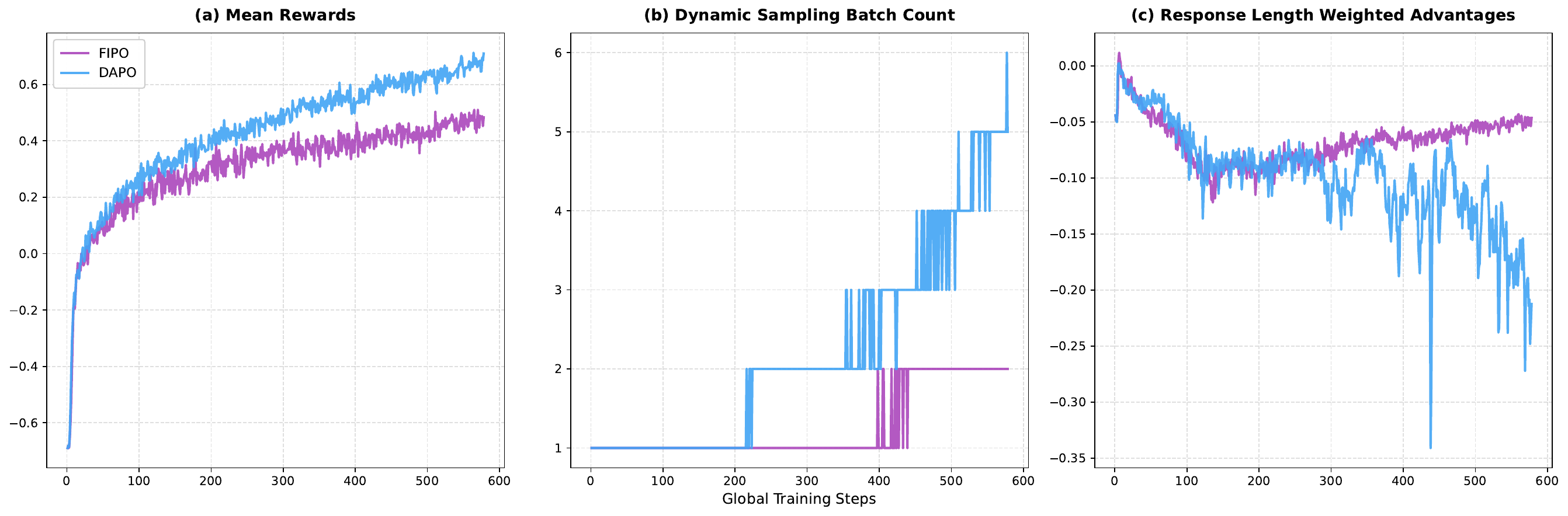}
    \caption{\textbf{Analysis of training reward and length-weighted advantages.}(a) \textbf{Mean training rewards.} DAPO achieves higher raw scores, which is expected as both methods incorporate an overlong penalty that suppresses the reward of longer responses. (b)\textbf{Number of Sampled Batches.} This metric indicates the sampling redundancy required to maintain a sufficient number of \textit{effective} batches. A higher sampling need suggests the model frequently generates non-informative trajectories on the training set, serving as a potential indicator of overfitting. (c) \textbf{Response length weighted mean advantages.} FIPO exhibits a sustained upward trend, establishing a positive reinforcement cycle where longer responses increasingly yield positive advantages. In contrast, DAPO shows a declining trend, suggesting a failure to convert length into effective reasoning gains, which ultimately limits its performance.}
    \label{fig:reward_adv}
    \vspace{-4mm}
\end{figure}

We further investigate the training dynamics by comparing the evolution of rewards and advantages. As shown in \autoref{fig:reward_adv}(a), the baseline (DAPO) consistently maintains a higher mean training reward than FIPO. However, we argue this disparity is a numerical artifact of the reward formulation rather than an indicator of superior performance. Because the reward function incorporates an overlong penalty, FIPO’s construction of elaborate reasoning chains inevitably leads to higher penalties, thus suppressing its average raw reward. Conversely, the baseline’s higher reward is driven by its tendency to generate shorter responses. While this strategy maximizes immediate reward by minimizing penalties, it suggests a convergence to a local optimum within a restricted search space. 

This hypothesis is further corroborated by the rapid escalation in the number of sampled batches for DAPO, as shown in \autoref{fig:reward_adv}(b). This trend indicates that the model is overfitting the training set, increasingly generating non-discriminative samples (i.e., batches that are uniformly correct or incorrect) which yield negligible gradient information. Consequently, the algorithm is forced to sample more aggressively to harvest sufficient effective data for optimization. In contrast, FIPO actively traverses a more expansive search space, prioritizing the structural depth required for challenging reasoning tasks over the mere avoidance of penalties.

This difference becomes even more pronounced when shifting from raw rewards to the dynamic incentives provided by advantages. As observed in \autoref{fig:reward_adv}(c), DAPO exhibits a declining trend in response length weighted mean relative advantage\footnote{We define the response length weighted mean advantage as: $\bar{A} = \frac{\sum_{i=1}^{B} \sum_{t=1}^{L_i} A_{i,t}}{\sum_{i=1}^{B} L_i}$, where $B$ is the batch size, $L_i$ is the response length of the $i$-th sample, and $A_{i,t}$ represents the token-level group relative advantage.} throughout training. This implies that the length of positive samples is increasingly dominated by that of negative samples, resulting in a diminishing incentive to extend derivations; since increased length no longer yields more positive relative advantages, the model eventually hits a plateau in reasoning growth. In stark contrast, FIPO demonstrates a consistent upward trajectory. This indicates that the positive samples are evolving to be significantly more substantive than their negative counterparts. \textbf{This dynamic fosters a sustained growth trajectory: as the generation of longer, valid reasoning chains yields increasingly positive advantages, facilitated by the steady rise in rewards, it preserves the model's momentum to pursue even more extensive and rigorous reasoning paths.}


\subsection{Smooth Policy Drift, Exploration and Gradient Update}

\begin{figure}[t] 
    \centering
    \includegraphics[width=\textwidth]{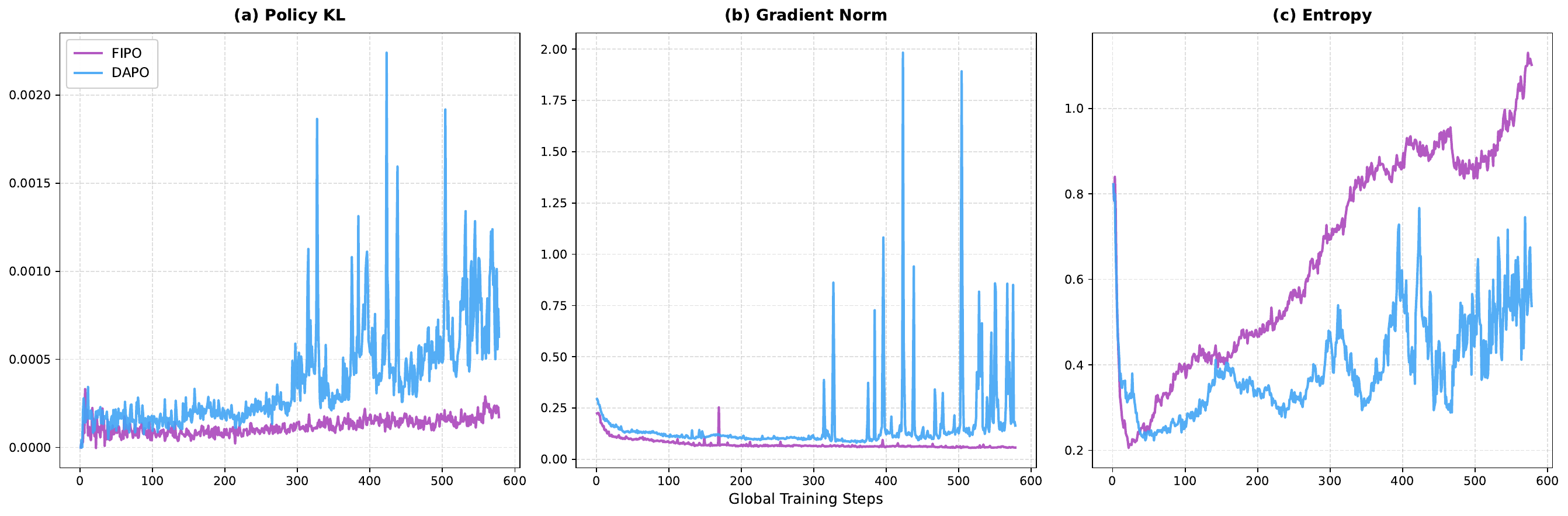}
    \caption{\textbf{Policy evolution and optimization dynamics.} (a) Policy KL divergence. (b) Policy Entropy. (c) Gradient Norm. FIPO exhibits a more controlled policy drift and smoother update gradients than DAPO. Notably, FIPO’s rising entropy, paired with the weighted advantage trends in Fig. 4(b), indicates that the model is actively exploring a broader reasoning space where longer CoT paths increasingly correspond to correct solutions.}
    \label{fig:other_metric}
    \vspace{-4mm}
\end{figure}

To further characterize the training process, we examine the evolution of policy behavior and optimization stability. As shown in \autoref{fig:other_metric}(a), FIPO exhibits a steady and structured increase in Policy KL divergence. This represents a progressive policy shift, where the model consistently moves away from its \textbf{previous policy state} to navigate toward a more specialized reasoning regime. This trend is qualitatively consistent with our rollout observations: the length of self-reflection segments increases incrementally rather than abruptly, reflecting a gradual expansion of the reasoning process (see Appendix \ref{app_sec: case_study} for examples).

The optimization characteristics also differ significantly in terms of gradient scale. As shown in \autoref{fig:other_metric}(b), FIPO’s Gradient Norm remains low and consistent throughout training, characterizing an evolution built upon fine-grained updates. In contrast, the baseline (DAPO) displays highly volatile fluctuations, with frequent, violent spikes in its gradient norm. These fluctuated updates indicate that DAPO's search process is at risk of abrupt shifts and potential instability.

This contrast in stability is further reflected in the policy entropy (\autoref{fig:other_metric}c). While FIPO maintains a smooth and sustained rise in entropy, indicating a continuous and stable exploration of the reasoning space, DAPO’s entropy is marked by noisy oscillations throughout the training process. In contrast, DAPO’s entropy is marked by noisy oscillations as training progressed. \textbf{Together, these metrics depict FIPO as a model that achieves significant and purposeful policy evolution toward complex reasoning while ensuring the optimization process remains numerically well-behaved.}

\section{Conclusion}

In this paper, we introduced \textbf{F}uture-KL \textbf{I}nfluenced \textbf{P}olicy \textbf{O}ptimization (\textbf{FIPO}), a reinforcement learning approach designed to resolve the coarse-grained credit assignment problem inherent in standard GRPO. By incorporating discounted \textit{Future-KL divergence} into policy updates, FIPO transforms sparse outcome-based rewards into dense, token-level supervision. Our empirical analysis identifies and addresses a critical ``length-performance plateau'' in existing baselines, demonstrating that standard uniform rewards fail to sustain long-chain reasoning. Validated on \textbf{Qwen2.5-32B-Base}, FIPO effectively breaks this ceiling: it propels performance on AIME 2024 from a baseline of 50.0\% to a \textbf{peak of 58.0\% (converging at 56.0\%)} and extends the average chain-of-thought length from 4,000 to over 10,000 tokens. Crucially, these findings challenge the prevailing assumption that complex critic models are necessary for granular credit assignment, proving that dense supervision can be effectively realized within the more efficient GRPO framework. To facilitate future research, we open-source our complete training code and recipes, providing the community with a scalable and accessible pathway to advance large-scale reasoning models.
\section{Limitations and Future Work}
 Despite its effectiveness, FIPO has certain limitations:
 
\textbf{Cost and Efficiency.} A primary constraint is the increased computational cost associated with extending reasoning sequences. As FIPO successfully unlocks CoT lengths exceeding 10,000 tokens, the training and inference overhead grows significantly, posing challenges for resource-constrained deployments. We argue that the development of advanced reasoning should be a sequential process: first eliciting long, high-quality reasoning capabilities, and subsequently optimizing them for efficiency. While this paper focuses on the first stage, breaking through length stagnation to achieve superior performance, the task of transforming these long reasoning paths into more concise and efficient forms is a critical next step. We will leave this for future exploration.

\textbf{Task Generalization.} Another limitation is that our evaluations are primarily conducted on mathematical reasoning benchmarks. However, we contend that mathematics serves as a rigorous and representative proxy for deep reasoning; its requirement for objective, verifiable ground truth and high-density logical consistency makes it the most demanding testbed for our algorithm. Having demonstrated that FIPO can overcome length stagnation in this challenging domain, we leave the exploration and validation of these elicited behaviors in other open-ended or less structured domains for future work.


\textbf{Training Data Scope.} To ensure a rigorous and fair comparison with the baseline, we restricted our training exclusively to the dataset used in DAPO. Consequently, we have not yet explored the scalability of FIPO on larger-scale or higher-quality datasets. While this controlled setting serves to isolate the algorithmic contributions of our method, the potential of FIPO when trained on more extensive or diverse data distributions remains uncharted. Moreover, while FIPO achieves superior performance over o1-mini on mathematical benchmarks, this advantage is inherently domain-specific. Given that our training was strictly confined to the math dataset, we do not anticipate these gains to generalize across non-mathematical domains, such as coding or symbolic logic, where o1-mini benefits from massive-scale, multi-stage reinforcement learning. Consequently, we leave the exploration of FIPO's generalization across broader data regimes and its fundamental scaling properties for future work. 

\textbf{Limited Model Scope.} A core objective of our study is to investigate RL-driven reasoning starting from a clean base model with no prior exposure to Long-CoT synthetic data. This strict requirement for experimental purity significantly limits the selection of suitable backbone models. Most contemporary open-source models optimized for reasoning have already undergone extensive supervised fine-tuning (SFT) or distillation from long-form reasoning traces. We contend that the underlying training dynamics of eliciting reasoning directly from a vanilla base model differ fundamentally from further optimizing a model that has already internalized distilled reasoning patterns. Consequently, our choice of models was restricted to a few high-quality vanilla base models, such as the Qwen2.5 series, to ensure that our findings specifically characterize the emergence of inherent reasoning potential rather than the refinement of pre-distilled CoT behaviors. In future work, we plan to investigate the efficacy and mechanistic behavior of our algorithm when applied to such pre-distilled Long-CoT models, exploring whether the dense advantage formulation can further refine or synergize with pre-existing distilled reasoning capabilities.

\textbf{Performance Gap vs. Distillation.} While RL-based self-evolution significantly enhances reasoning, it remains a "discovery-based" process that is inherently less efficient than direct distillation. Larger teacher models provide a much denser supervisory signal and superior heuristics (logits) that are difficult for a smaller model to self-derive through sparse rewards alone, resulting in a persistent performance gap between self-trained and distilled variants.
\clearpage
\section{Contributions}

\noindent \textcolor{mypurple}{\textbf{Core Contributors}} \\
Chiyu Ma$^{1,5}$, Shuo Yang$^{2,5}$

\noindent \textcolor{mypurple}{\textbf{Contributors}} \\
Kexin Huang$^{5}$, Jinda Lu$^{5}$, Haoming Meng$^{3,5}$, Shangshang Wang$^{4,5}$

\vspace{0.3cm}

\noindent \textcolor{mypurple}{\textbf{Supervision}} \\
Bolin Ding$^{6}$, Soroush Vosoughi$^{1}$, Guoyin Wang$^{5}$, Jingren Zhou$^{6}$

\vspace{0.3cm}

\noindent \textcolor{mypurple}{\textbf{Affiliations}} \\
$^{1}$ Dartmouth College \\
$^{2}$ Peking University \\
$^{3}$ University of Toronto \\
$^{4}$ University of Southern California \\
$^{5}$ Qwen Pilot Team \\
$^{6}$ Alibaba

\clearpage
\bibliography{biblio}
\bibliographystyle{colm2024_conference}

\newpage
\clearpage
\appendix
\startcontents[appendices]
\section*{Appendix Table of Contents}
\printcontents[appendices]{}{1}[2]{\large}
\newpage

\section{Parameter Settings}
\label{sec: parameters}
We detail the specific hyperparameter configurations employed for fine-tuning Qwen2.5-32B-Base and Qwen2.5-7B-Math. We started from the publicly released DAPO training srcipt. To ensure full reproducibility, we will release our complete codebase along with the training scripts used in our experiments.

\subsection{Qwen2.5 32B Base}

We conduct training using a global batch size of 512 with a group size of $G=16$ generations per prompt. The model is optimized using a learning rate of $1\times 10^{-6}$ and a weight decay of 0.1. For the FIPO specific parameters, we set the Future-KL decay rate to 32.0 and employ a safety threshold of 10.0 to filter extreme influence weights. The policy updates are constrained with asymmetric clipping ratios of $[0.2, 0.28]$. To support extensive reasoning chains, the maximum response length is set to 20,480 tokens. A detailed hyperparameter comparison between our method and DAPO can be found in \autoref{tab:hyperparameters_32b}. 

\begin{table}[h]
    \centering
    \small
    \caption{\textbf{Hyperparameter settings for Qwen2.5-32B-Base experiments.} We compare the configuration of our proposed FIPO against the DAPO baseline. Most infrastructure and optimization settings remain identical to ensure a fair comparison.}
    \label{tab:hyperparameters_32b}
    \begin{tabular}{l|c|c}
        \toprule
        \textbf{Hyperparameter} & \textbf{DAPO (Baseline)} & \textbf{FIPO (Ours)} \\
        \midrule
        \multicolumn{3}{c}{\textit{Shared Optimization Settings}} \\
        \midrule
        Base Model & \multicolumn{2}{c}{Qwen2.5-32B-Base} \\
        Global Batch Size & \multicolumn{2}{c}{512} \\
        Group Size ($G$) & \multicolumn{2}{c}{16} \\
        Learning Rate & \multicolumn{2}{c}{1e-6} \\
        LR Scheduler & \multicolumn{2}{c}{Constant with 10 Warmup Steps} \\
        Weight Decay & \multicolumn{2}{c}{0.1} \\
        Gradient Clipping & \multicolumn{2}{c}{1.0} \\
        Max Prompt Length & \multicolumn{2}{c}{2,048} \\
        Max Response Length & \multicolumn{2}{c}{20,480} \\
        Overlong Buffer & \multicolumn{2}{c}{4096} \\
        Sampling Temp / Top-p & \multicolumn{2}{c}{1.0 / 1.0} \\
        Dual Clip Ratio  & \multicolumn{2}{c} {10.0} \\
        Policy Clip Ratio & \multicolumn{2}{c}{[0.2, 0.28] (Asymmetric)} \\
        KL Penalty Coef & \multicolumn{2}{c}{0.0} \\
        \midrule
        \multicolumn{3}{c}{\textit{Method-Specific Settings}} \\
        \midrule
        Mini-Batch Size & 32 & 64 \\
        Loss Function & DAPO & \textbf{Future-KL} \\
        Future-KL Decay Rate & - & 32.0 \\
        Future-KL Clip Ratio & - & [1.0, 1.2] \\
        Safety Threshold & - & 10.0 \\
        \bottomrule
    \end{tabular}
\end{table}

\subsection{Qwen2.5 7B Math}

We initially validated the effectiveness of our approach on Qwen2.5-7B-Math as a pilot study before scaling to the 32B parameter regime. During this preliminary phase, we observed that training performance was initially volatile, and the reproducibility of reasoning gains was inconsistent across independent runs. To address these stability issues, we increased the group size to $G=32$ to provide a more stable optimization signal and enforced a stricter advantage clipping threshold of $3.0$ to filter out destabilizing updates. \textbf{These adjustments successfully stabilized the training trajectory, making the results more reliable while maintaining the same performance level.} The model was optimized using a learning rate of $1\times 10^{-6}$ and a weight decay of $0.1$, with the Future-KL decay rate set to $32.0$. Following the stabilization of these pilot experiments, we scaled the validated framework to the 32B model. A summary of the 7B hyperparameter settings is provided in \autoref{tab:hyperparameters_7b}, and detailed results of the 7B MATH experiments are available in Sec. \ref{app_sec: 7B_result}.

\begin{table}[h]
    \centering
    \small
    \caption{\textbf{Hyperparameter settings for Qwen2.5-7B-MATH experiments.} We compare the configuration of our proposed FIPO against the DAPO baseline. Most infrastructure and optimization settings remain identical to ensure a fair comparison.}
    \label{tab:hyperparameters_7b}
    \begin{tabular}{l|c|c}
        \toprule
        \textbf{Hyperparameter} & \textbf{DAPO (Baseline)} & \textbf{FIPO (Ours)} \\
        \midrule
        \multicolumn{3}{c}{\textit{Shared Optimization Settings}} \\
        \midrule
        Base Model & \multicolumn{2}{c}{Qwen2.5-7B-MATH} \\
        Global Batch Size & \multicolumn{2}{c}{512} \\
        Learning Rate & \multicolumn{2}{c}{1e-6} \\
        LR Scheduler & \multicolumn{2}{c}{Constant with 10 Warmup Steps} \\
        Weight Decay & \multicolumn{2}{c}{0.1} \\
        Gradient Clipping & \multicolumn{2}{c}{1.0} \\
        Max Prompt Length & \multicolumn{2}{c}{2,048} \\
        Sampling Temp / Top-p & \multicolumn{2}{c}{1.0 / 1.0} \\
        Dual Clip Ratio  & \multicolumn{2}{c} {10.0} \\
        Policy Clip Ratio & \multicolumn{2}{c}{[0.2, 0.28] (Asymmetric)} \\
        KL Penalty Coef & \multicolumn{2}{c}{0.0} \\
        \midrule
        \multicolumn{3}{c}{\textit{Method-Specific Settings}} \\
        \midrule
        Mini-Batch Size & 32 & 64 \\
        Loss Function & DAPO & \textbf{Future-KL} \\
        Future-KL Decay Rate & - & 32.0 \\
        Future-KL Clip Ratio & - & [0.8, 1.2] \\
        Safety Threshold & - & 3.0 \\
        Group Size ($G$) & 16 & 32 \\
        Max Response Length & 8192 & 10240 \\
        Overlong Buffer & 4096 & 2048 \\
        \bottomrule
    \end{tabular}
    \vspace{-3mm}
\end{table}


\section{Result on Qwen2.5 7B Math}
\label{app_sec: 7B_result}
This section documents the experimental results on the Qwen-2.5 7B MATH model. Due to the prohibitive computational cost of 32B models, we conducted the majority of our early-stage explorations and ablations using this variant. As noted in Section 3.2, the 7B model exhibited noticeable performance sensitivity across different training configurations. To address this, we performed targeted hyperparameter adjustments to stabilize the training dynamics and ensure the reliability of the observed trends. \textit{Note that we extend the context window from 4096 to 32768, following the instructions from DAPO verl scripts.}

\subsection{Performance}

\autoref{tab:7b_perf} presents a comparative analysis of Pass@1 performance across different RL-based methods on the AIME 2024 and AIME 2025 benchmarks. Our proposed FIPO (7B) achieves a notable performance of 40.0\% on AIME 2024, significantly outperforming the GRPO (7B) baseline (22.0\%) and the DAPO (7B) method (36.0\%). However, we observe a general performance compression on the AIME 2025 benchmark. While FIPO still maintains a leading edge at 19.0\%, the performance gap between the three methods narrows considerably compared to the AIME 2024 results. This phenomenon is primarily attributed to the substantially higher intrinsic difficulty and the "live" nature of the AIME 2025 problems, which appear to approach the reasoning ceiling of 7B-parameter models without external guidance. At this scale, especially in the absence of pre-training on more advanced chain-of-thought structures, the increased complexity of the reasoning chains required for AIME 2025 makes it challenging to distinguish the marginal gains of different optimization algorithms, as most models encounter similar structural bottlenecks.

\begin{table}[ht]
\centering
\caption{\textbf{Qwen 2.5-7B-MATH Performance Comparison of Pass@1 Performance on AIME2024 and AIME2025.} All results are reported as percentages (\%). We report the peak average Pass@1 across 32 samples (Avg@32). To reduce sensitivity to digit-level generation variance, final values are rounded to the nearest integer.}
\label{tab:7b_perf}
\begin{tabular}{lcc}
\toprule
\textbf{Method} & \textbf{AIME 2024 (Pass@1)} & \textbf{AIME 2025 (Pass@1)} \\ 
\midrule
GRPO \citep{guo2025deepseek} & 22.0\% & 18.0\% \\
DAPO & 36.0\% & 18.0\% \\
\textbf{FIPO (Ours)}& \textbf{40.0\%}& \textbf{19.0\%} \\
\bottomrule
\end{tabular}
\end{table}

\subsection{Result Analysis}

\begin{figure}[ht] 
    \centering
    \includegraphics[width=\textwidth]{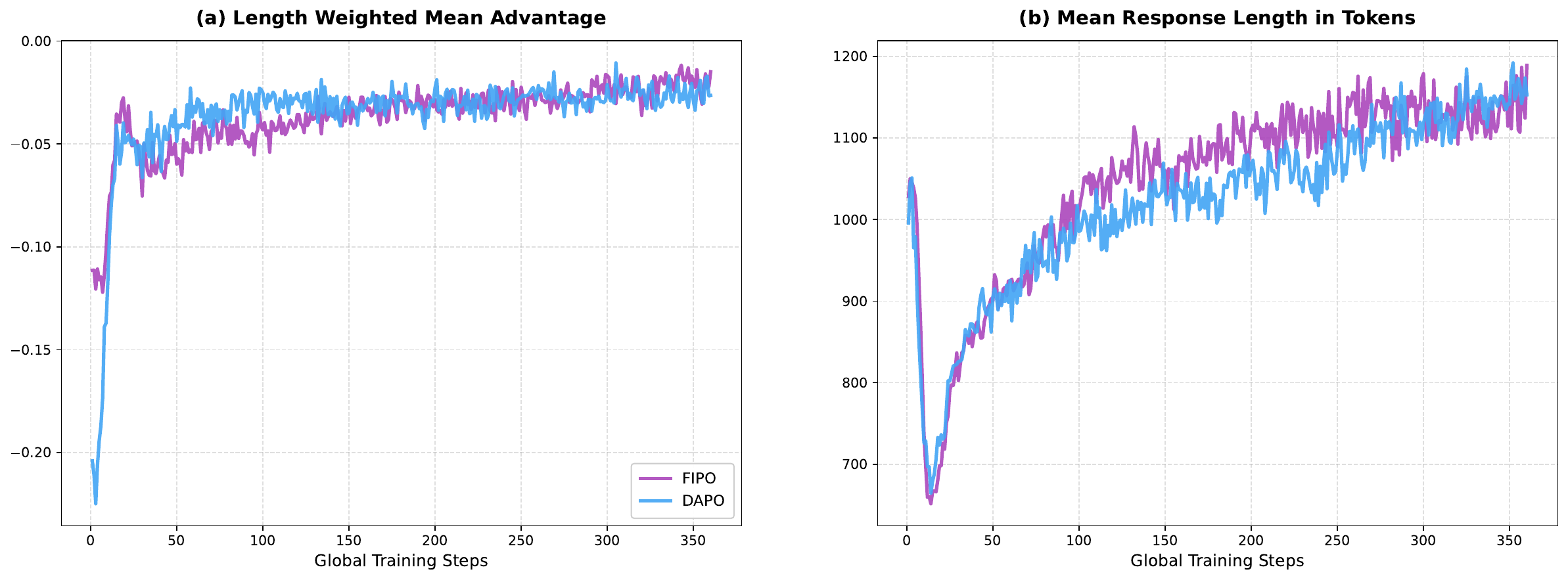}
    \caption{\textbf{Analysis of length-weighted advantages and mean response length of Qwen2.5-7B-MATH.} (a) Subfigure (a) shows FIPO maintains a stable Length-Weighted Mean Advantage. (b) Subfigure (b) shows the mean response length remains steady around 1200 tokens. This suppressed length growth is likely caused by the 4K context window limit of the base model, and its inherent bias toward concise code-based reasoning, which anchors the generation depth.} 
    \label{fig:reward_adv_7b}
    \vspace{-4mm}
\end{figure}

As shown in \autoref{fig:reward_adv_7b}, both algorithms maintain a stable length-weighted mean advantage and a mean response length fluctuating around 1200 tokens. Notably, this stands in sharp contrast to the 32B model training, where FIPO typically triggers a sustained growth in response length. We hypothesize that this length stagnation is not an optimization failure, but rather a reflection of the 7B model's inherent capacity limits and training priors. Specifically, the Qwen-2.5 7B MATH base was pre-trained with a restricted 4K context window \citep{yang2024qwen2}, which likely imposes a physical ceiling on its reasoning depth without external guidance. Furthermore, the model's strong initial bias toward code-based reasoning \citep{shao2025spurious}, which favors logically dense, deterministic pathways over verbose exploration, combined with potential AIME24 data leakage \citep{wu2025reasoning}, provides \textbf{"high-confidence shortcuts"} that allow the model to reach concise solutions quickly rather than exploring longer, iterative paths.

\begin{figure}[ht] 
    \centering
    \includegraphics[width=\textwidth]{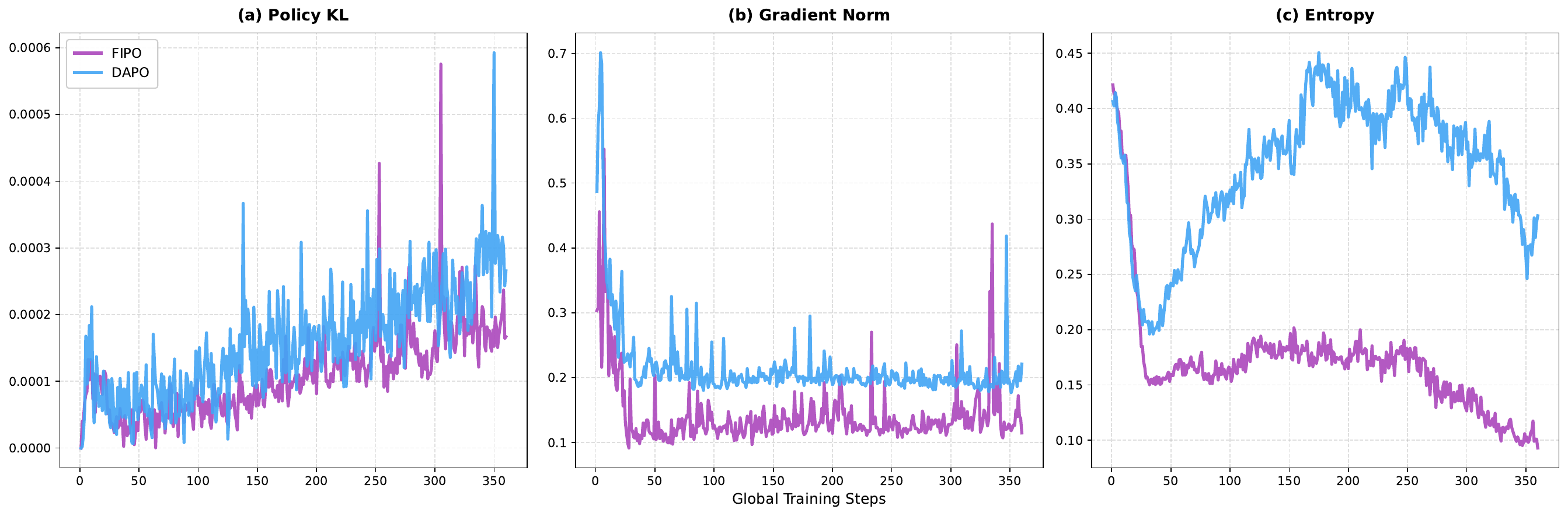}
    \caption{\textbf{Policy evolution and optimization dynamics of Qwen2.5-7B-MATH.} (a) Policy KL and (b) Gradient Norm illustrate the optimization path. (c) Notably, FIPO results in lower entropy compared to DAPO, indicating that the model converges toward more deterministic reasoning traces during the RL process.}    \label{fig:other_metric_7b}
    \vspace{-4mm}
\end{figure}

This phenomenon of "efficient but constrained" reasoning is further corroborated by the entropy dynamics shown in \autoref{fig:other_metric_7b}(c). While larger models typically rely on sustained entropy growth to explore complex reasoning spaces, the 7B MATH model achieves optimal performance by converging onto a markedly lower entropy policy under FIPO. This divergence suggests a fundamental difference in scaling behavior: while the 32B model benefits from broad exploration, the 7B model appears to prioritize the refinement of a specific, high-confidence reasoning manifold.

The hypothesis that a low-entropy state is critical for performance at this scale is further supported by our ablation study on the Future-KL influence weight clipping. As discussed in \autoref{sec: ablation}, the 7B and 32B models exhibit different sensitivities to the Future-KL clipping range. While the 32B model maintains superior performance with an influence weight clipped between 1.0 and 1.2, applying this same range to the 7B model leads to a continuous increase in entropy, similar to the behavior of the 32B model, yet results in a significant degradation in performance. Instead, we found that the 7B model performs optimally under a different regime where a clipping range of 0.8 to 1.2 is applied. This observation suggests that the 7B model may lack sufficient inherent self-exploration capacity to derive benefit from higher-entropy states. \textbf{In this regime, maintaining higher entropy appears to introduce more detrimental noise than useful discovery. Consequently, at this scale, superior performance is most likely attainable when the model converges to specific, low-entropy reasoning traces. This phenomenon is fundamentally consistent with the principles of entropy minimization \citep{agarwal2025unreasonable}, entropy regulation \citep{wu2025quantile}, self-guiding \citep{zuo2025ttrl}, and self-certainty optimization \citep{zhao2025learning}, which have typically been observed to be effective in models of similar scale.}

\section{Ablation Studies}
\label{sec: ablation}

In this section, we provide a comprehensive analysis of the ablation studies conducted to validate our methodology. We first examine the impact of high-value clipping and the extension of the maximum response length on the resulting mean response length during 32B model training with FIPO. Next, we present results on the 7B model scale regarding weight filtration, which complement the stability-related findings observed in our 32B experiments. Furthermore, we investigate the implications of various adaptive clipping configurations and their influence on performance, specifically highlighting how these choices drive divergent behaviors in policy entropy. Lastly, we evaluate the sensitivity of the optimization process to different decay rate options and their potential impact on convergence dynamics.

\subsection{Clip-High, Max Length, and Response Length}
\label{sec:cliphigh_ablation}
\begin{figure}[ht] 
    \centering
    \includegraphics[width=\textwidth]{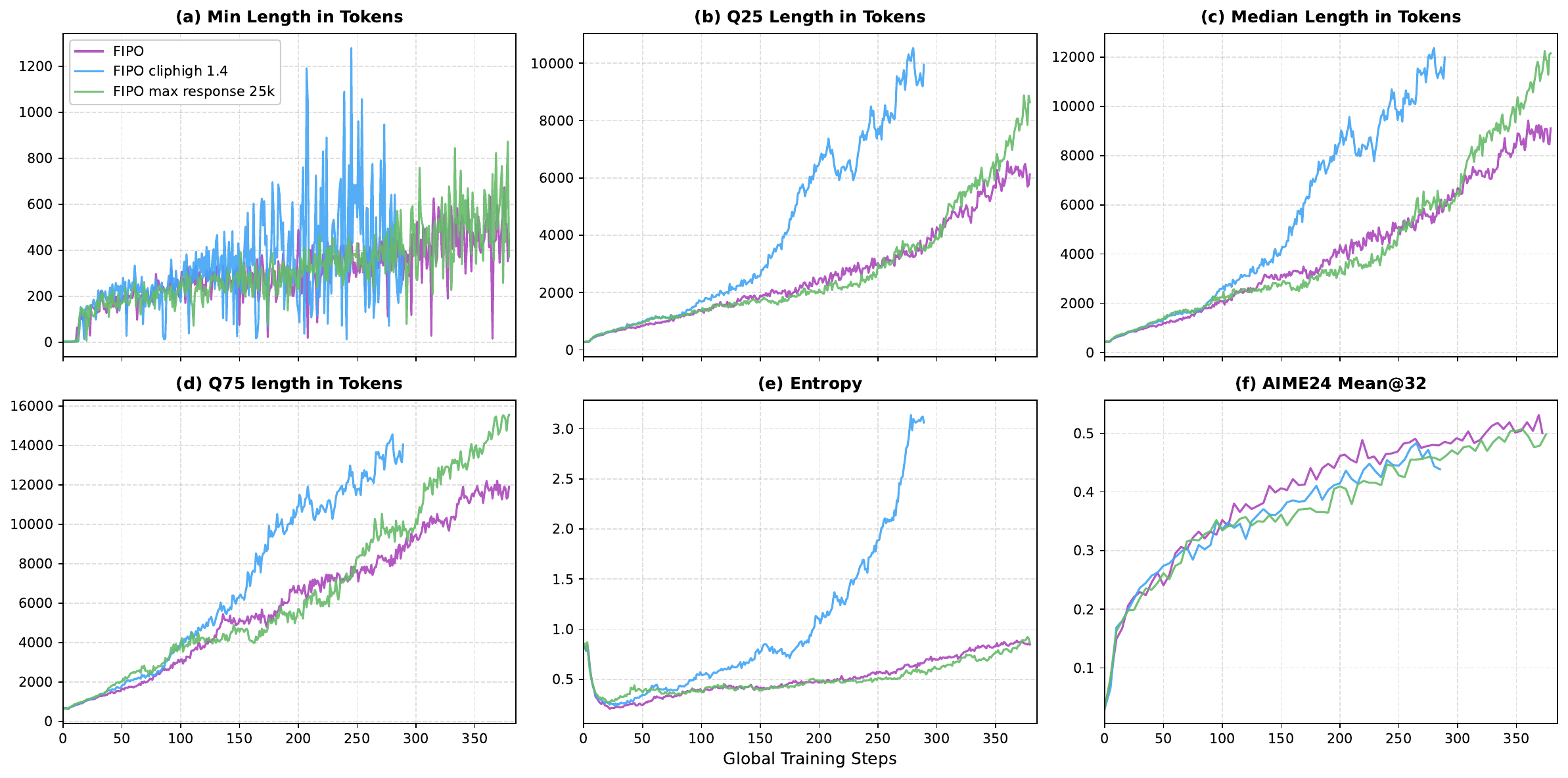}
    \caption{\textbf{Ablation study on Response Length Control on Qwen2.5-32B-Base.} We compare the baseline FIPO with FIPO under cliphigh 1.4 and FIPO with a max resonse length of 25K. (a-d) Increasing the maximum length limit triggers a significant early surge in response length, yet results in lower training efficiency and marginal performance gains. (e) A higher clip ratio causes a sharp escalation in policy entropy, indicating potential training instability. (f) Overall, while both modifications encourage longer responses, they are less effective in improving AIME24 performance compared to the balanced FIPO configuration.}
    \label{fig:ablation_length}
    \vspace{-3mm}
\end{figure}

One of the most interesting phenomena we observed is an unexpected surge in response length at the early stage when turning up the clip-high ratio $\epsilon_{high}$ of the PPO objective. In this study, we set $\epsilon_{high}=1.4$, providing a larger trust region for policy updates when the advantage is positive. For the maximum length trial, we extended the limit to 25K tokens (from the default 20K) while maintaining the overlong buffer at a consistent 20\% ratio. All other variables remained consistent to ensure a fair comparison. As shown in Fig. \autoref{fig:ablation_length}(a-d), a higher $\epsilon_{high}$ allows the policy to deviate more aggressively from the old distribution when receiving positive advantages, incentivizing the model to "over-explore" longer reasoning paths almost immediately. However, this surge comes at the cost of training stability; as depicted in \autoref{fig:ablation_length}(e), the policy entropy explodes under a higher $\epsilon_{high}$, indicating that the optimization landscape becomes volatile when the clipping constraint is too relaxed.

A strikingly similar failure mode is observed when increasing the maximum response length limit. While this also triggers a rapid growth in token count, it reveals a clear diminishing return in reasoning efficiency. Both cases suffer from the emergence of superficial reasoning behaviors, such as repetitive content, task-irrelevant LaTeX formatting, and premature self-reflection. We contend that introducing self-reflection can be counterproductive when the model has not yet established a solid foundation for one-step reasoning. In such cases, the model lacks the verifiable internal logic needed to evaluate its own state, leading to internal consistency conflicts, a phenomenon where the model oscillates between conflicting steps without progressing toward the correct solution. \textbf{These observations suggest that the emergence of higher-order reasoning patterns, such as self-reflection, is not necessarily "the earlier, the better" during the training process.} Instead, it should be a progressive maturation where complex strategies build upon a stable base of fundamental logic. Consequently, the performance gain on AIME24 (\autoref{fig:ablation_length}f) is marginal compared to the balanced FIPO baseline. \textbf{These results underscore the importance of governed growth in response length, ensuring that extended computation is allocated to genuine logical depth rather than redundant self-correction cycles.}

\subsection{Extreme Value Filtering}

As we noted in \autoref{sec:future_kl}, since the computation of Future-KL is highly dependent on the logit values between each update, one of its drawbacks is its sensitivity to importance ratio volatility. If the variation in importance ratios becomes significant, the Future-KL weight can become highly volatile at the same time. Rather than providing a stable guiding signal for policy updates, the excessive noise in importance ratios causes the Future-KL term to lose its effectiveness. In such cases, the incorporation of future KL information could result in less stable updates, and the model's performance tends to revert to a suboptimal level close to the standard baseline without showing the intended improvements. This is particularly observable in 7B models, which exhibit frequent fluctuations in importance ratios.

\autoref{tab:7b_filter} presents the ablation results on the 7B model. We observe that without filtering extreme IS ratios, the performance gain on AIME2024 is suboptimal compared to the configuration where filtering is applied. Although the unfiltered version achieves a slightly higher score on AIME2025 in one instance, the overall reliablity and the peak performance on AIME2024 (40.0\%) along with the observations made on 32B training confirm the necessity of the filtering mechanism for a more reliable training. 

\begin{figure}[t] 
    \centering
    \includegraphics[width=\textwidth]{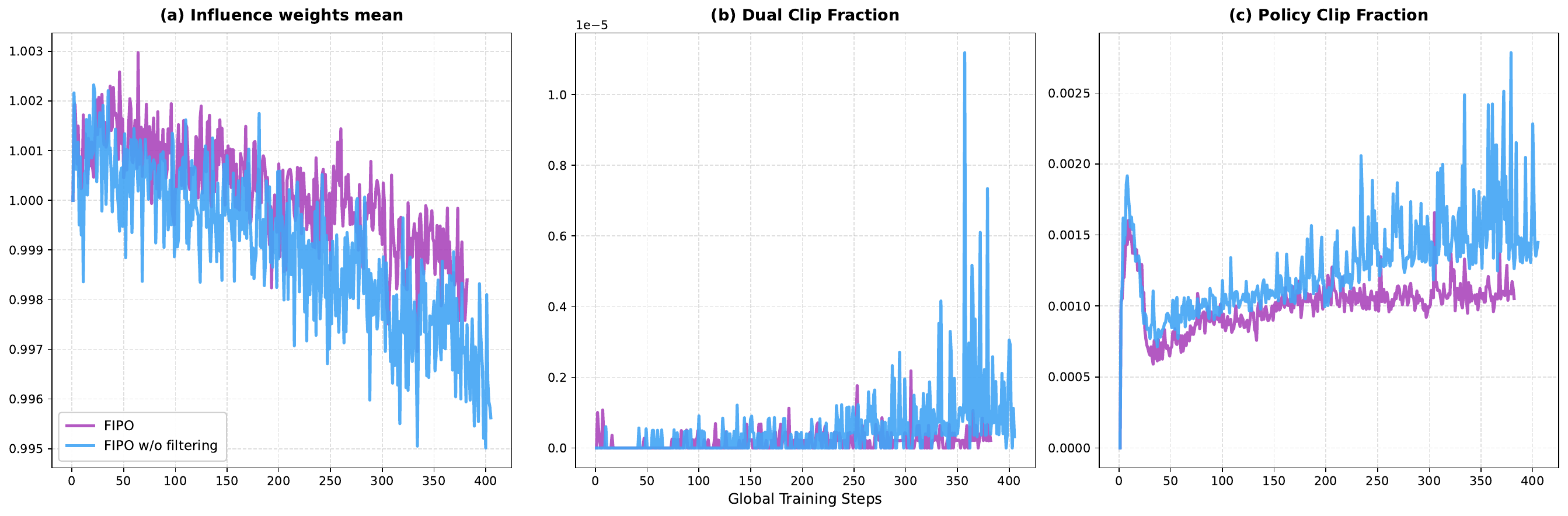}
    \caption{\textbf{Ablation Study on Filtering Mechanism in Future-KL on Qwen2.5-7B-MATH.} 
    We compare the training stability of FIPO with and without filtering extreme importance sampling (IS) ratios. 
    (a) Future-KL influence weights: removing extreme IS ratios results in a significantly more constrained and compact weight range, leading to more stable policy updates. 
    (b-c) Clipping statistics: extreme IS ratios rise disproportionate influence weights that drive more aggressive updates. This is reflected in the increased dual clip and policy clip fractions, where a larger proportion of tokens are pushed beyond the trust region when filtering is absent.}
    \label{fig:ablation_filter}
    \vspace{-3mm}
\end{figure}

Moreover, as shown in \autoref{fig:ablation_filter}(a), removing extreme IS ratios results in a significantly more constrained and compact weight range, which provides a more stable signal for the advantage estimator. In contrast, the absence of filtering leads to disproportionately large influence weights. This leads to a higher frequency of aggressive policy updates, as evidenced by the clipping fractions in \autoref{fig:ablation_filter}(b) and (c). When filtering is absent, a larger proportion of tokens are pushed beyond the designated trust region, resulting in higher dual clip and policy clip fractions. This suggests that the model is frequently attempting to make updates that exceed the constraints of the PPO objective, thereby explaining the suboptimal gains in performance when IS ratio volatility is high.

\begin{table}[ht]
\centering
\caption{\textbf{Ablation on Extreme Value Filtering and Influence Weight Clipping Performance Comparison of Pass@1 Performance on AIME2024 and AIME2025 on Qwen2.5-7B-MATH.} All results are reported as percentages (\%). We report the peak average Pass@1 across 32 samples (Avg@32).}
\label{tab:7b_filter}
\begin{tabular}{lcc}
\toprule
\textbf{Method} & \textbf{AIME 2024 (Pass@1)} & \textbf{AIME 2025 (Pass@1)} \\ 
\midrule
FIPO ($\epsilon_{f_{low}}=1.0$, $\epsilon_{f_{high}}=1.2$) & 36.0\%  & 19.0\% \\
FIPO (w/o filtering, $\epsilon_{f_{low}}=0.8$, $\epsilon_{f_{high}}=1.2$)& 38.0 \% & \textbf{21.0}\% \\
\textbf{FIPO ($\epsilon_{f_{low}}=0.8$, $\epsilon_{f_{high}}=1.2$)}& \textbf{40.0\%}& 19.0\% \\
\bottomrule
\end{tabular}
\end{table}

\subsection{Influence Weight Clipping}
\begin{figure}[t] 
    \centering
    \includegraphics[width=0.8\textwidth]{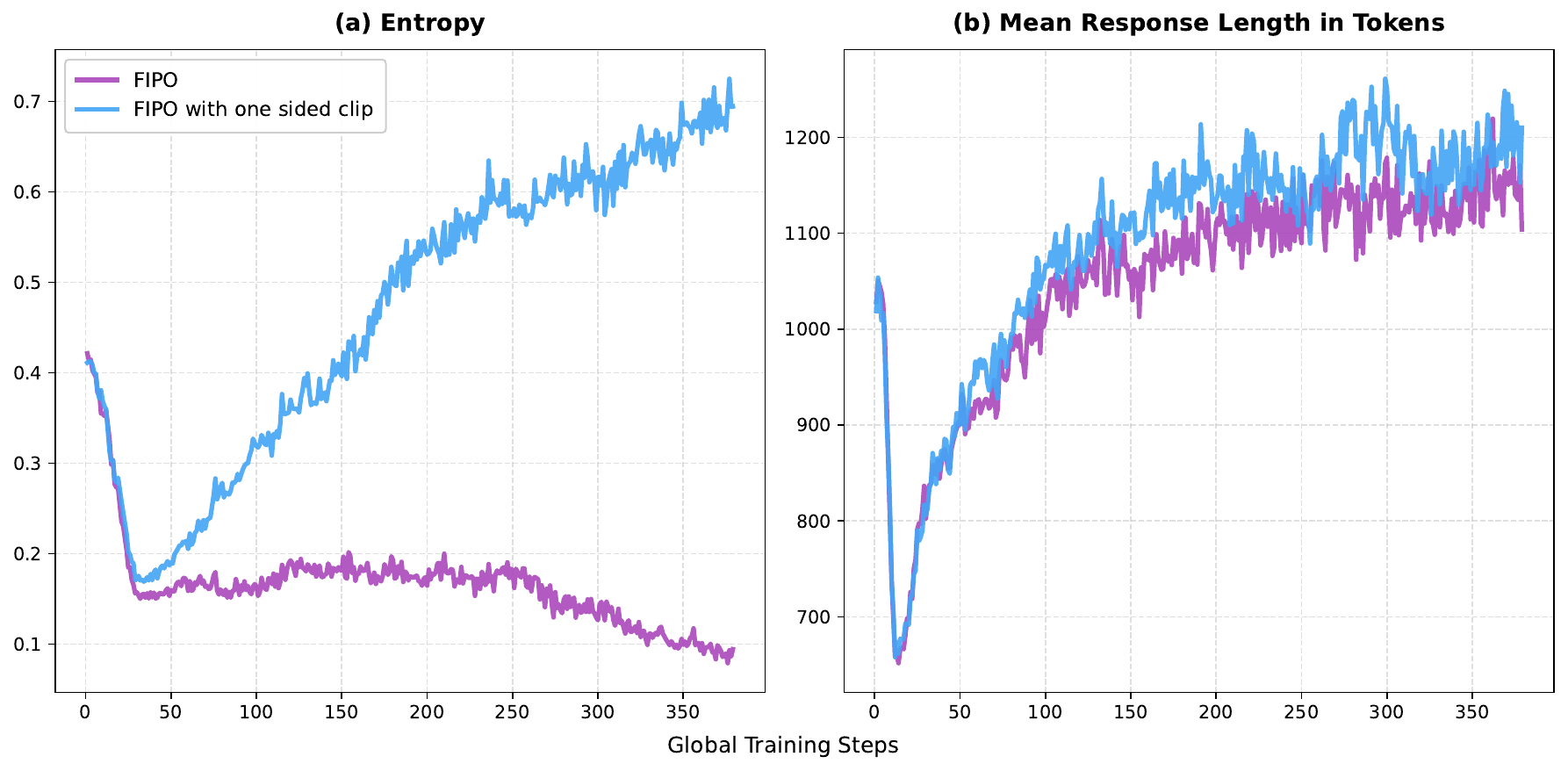}
    \caption{\textbf{Ablation Study on Influence Weight Clipping on Qwen-2.5-7B-MATH.} 
    (a) Similar to 32B, a influence weight with range $[1.0, 1.2]$ resulted in a continuous growth in entropy, thus encouraging more explorations.
    (b) However, we do not observe any surge in response length as we discussed in the previous section.}
    \label{fig:ablation_iwc}
    \vspace{-3mm}
\end{figure}

\autoref{tab:7b_filter} shows the result of ablations on the influence weight clipping. We primarily modified the clipping range with respect to 1. An influence weight with a range of $[1.0, 1.2]$ essentially provides more penalty for negative samples and more reward to positive ones. The range of $[0.8, 1.2]$ inherits these properties while offering a more balanced influence; it further reduces the reward when a token in a positive sample is associated with subsequent negative behaviors, and reduces the penalty when a token in a negative sample is associated with subsequent positive behaviors. This helps to provide a more controlled exploration and thus resulted in the performance improvements shown in \autoref{tab:7b_filter}.

Specifically, the $[0.8, 1.2]$ configuration achieves $40.0\%$ on AIME 2024, surpassing the $36.0\%$ of the $[1.0, 1.2]$ setting. While the latter encourages more aggressive exploration, as evidenced by the continuous growth in policy entropy shown in \autoref{fig:ablation_iwc}(a), our results suggest that 7B models are sensitive to such excessive exploration pressure. This aligns with our broader observation on scaling behavior: unlike larger models that benefit from broad exploration, the 7B model appears to prioritize the refinement of a specific, high-confidence reasoning manifold. In this context, the higher entropy induced by the $[1.0, 1.2]$ range introduces more detrimental noise than useful discovery. By contrast, the balanced influence of $[0.8, 1.2]$ facilitates convergence toward a lower-entropy state, optimizing for self-certainty rather than stochastic search.

\subsection{Effective Horizon of Decay Rate}
\begin{figure}[t] 
    \centering
    \includegraphics[width=\textwidth]{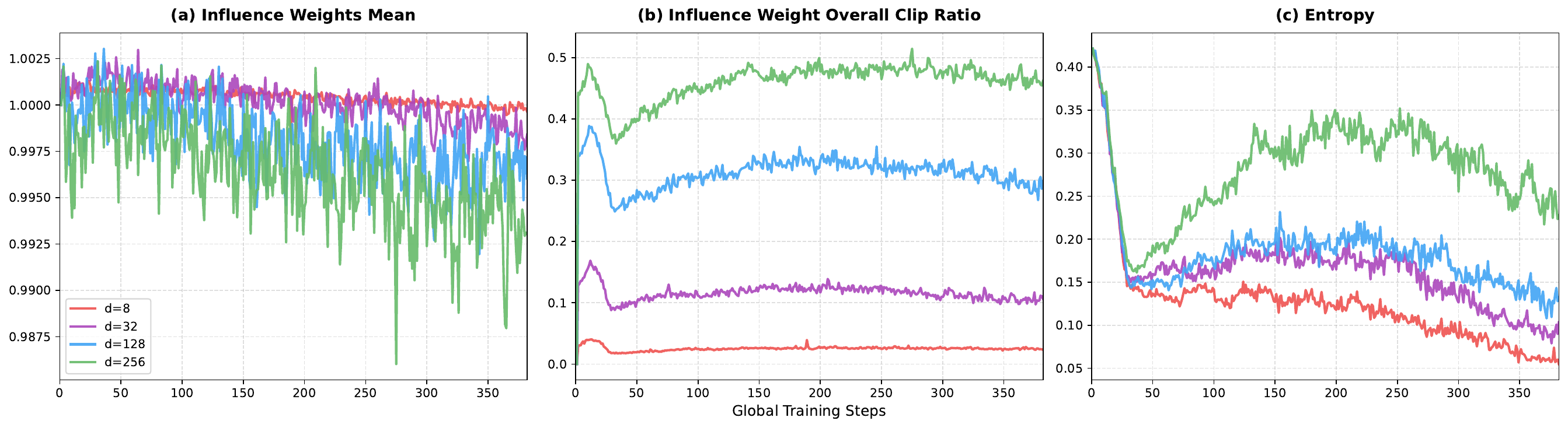}
    \caption{\textbf{Ablation Study on the Decay Rate Horizon ($\tau$) on Qwen2.5-7B-MATH.} 
    (a) Mean Influence Weight: Extending the decay horizon (e.g., $\tau=256$) significantly increases weight fluctuations, potentially destabilizing policy updates. Conversely, a highly restrictive horizon (e.g., $\tau=8$) limits the weights too close to the baseline. 
    (b) Influence Weight Clip Ratio: A larger decay horizon results in a higher clipping frequency. This reflects more drastic variations in the raw influence weights, directly corroborating the trends observed in (a).
    (c) Entropy: An intermediate horizon ($\tau=32$) strikes a critical balance. It avoids the excessive volatility induced by larger decay rates, as well as the premature entropy collapse caused by the myopic guidance of shorter horizons, allowing the 7B model to maintain controlled and effective exploration.}
    \label{fig:decay_ablation}
    \vspace{-3mm}
\end{figure}
To complement our analysis, we further perform an ablation study on the effective horizon of the decay rate, specifically testing values of $\tau \in \{8, 32, 128, 256\}$. These values define the half-life of the influence weight decay, determining the token-distance over which future KL information remains significant. \autoref{tab:decay_ablation_perf} presents the performance across these configurations, while \autoref{fig:decay_ablation} illustrates the corresponding training dynamics. As shown in \autoref{fig:decay_ablation}(a), the magnitude of fluctuation in Mean Influence Weight increases as the decay horizon $\tau$ is extended. For the largest horizon ($\tau=256$), the weight mean exhibits the most significant deviation from 1.0, which can introduce instability into the policy updates. In contrast, for a very small horizon ($\tau=8$), the influence weight remains highly proximal to 1.0, effectively causing the model to stay close to the baseline and relatively underutilize future KL information.

The entropy dynamics in \autoref{fig:decay_ablation}(c) further corroborate our earlier observations regarding the exploration capacity of the 7B model. While all configurations follow a characteristic trend of an initial decrease followed by a rise and eventual decline, the overall entropy levels are markedly higher for larger decay rates. The $\tau=256$ setting maintains the highest entropy throughout most of the training, indicating aggressive exploration driven by long-range future signals. Just as we observed with the restrictive $[1.0, 1.2]$ influence weight range, this sustained high-entropy state introduces excessive volatility that the 7B model struggles to accommodate. Conversely, the rapid entropy drop observed in the $\tau=8$ setting represents an entirely different failure mode: because its influence weights fluctuate only marginally, the model receives highly myopic guidance. Lacking sufficient long-term foresight, the policy prematurely collapses into a suboptimal, low-entropy state. The intermediate horizon, $\tau=32$, strikes a critical balance. Similar to the stabilizing effect of the $[0.8, 1.2]$ clipping range, $\tau=32$ provides enough localized future signal to safely navigate the updates without causing premature stagnation, allowing the policy to ultimately converge onto a specific, high-quality reasoning manifold. This alignment across different hyperparameters solidifies our hypothesis: \textbf{at the smaller scale such as 7B, optimizing for self-certainty through controlled, moderate exploration is more effective than forcing broad, high-entropy searches.
}
\begin{table}[ht]
\centering
\vspace{-3mm}
\caption{\textbf{Ablation on the Effective Horizon of Decay Rate Performance Comparison of Pass@1 Performance on AIME2024 and AIME2025 on Qwen2.5-7B-MATH.} All results are reported as percentages (\%). We report the peak average Pass@1 across 32 samples (Avg@32).}
\label{tab:decay_ablation_perf}
\begin{tabular}{lcc}
\toprule
\textbf{Method} & \textbf{AIME 2024 (Pass@1)} & \textbf{AIME 2025 (Pass@1)} \\ 
\midrule
FIPO ($\tau$=8)& 40.0\%& 17.0\% \\
\textbf{FIPO ($\tau$=32)}& 40.0\%& 19.0\% \\
FIPO ($\tau$=128)& 39.0\%& \textbf{21.0\%} \\
FIPO ($\tau$=256)& \textbf{42.0\%}& 16.0\% \\

\bottomrule
\end{tabular}
\end{table}

\section{Case Study}
\label{app_sec: case_study}

In this section, we provide a qualitative analysis of the model's reasoning evolution by examining specific outputs on the AIME 2024 competition. We randomly selected the responses from the grouped samples. The transition from DAPO's length stagnation to FIPO's sustained scaling is not merely a quantitative change in token count, but a qualitative transformation in how the model utilizes its "thinking" budget. 

\begin{figure}[t] 
    \centering
    \includegraphics[width=\textwidth]{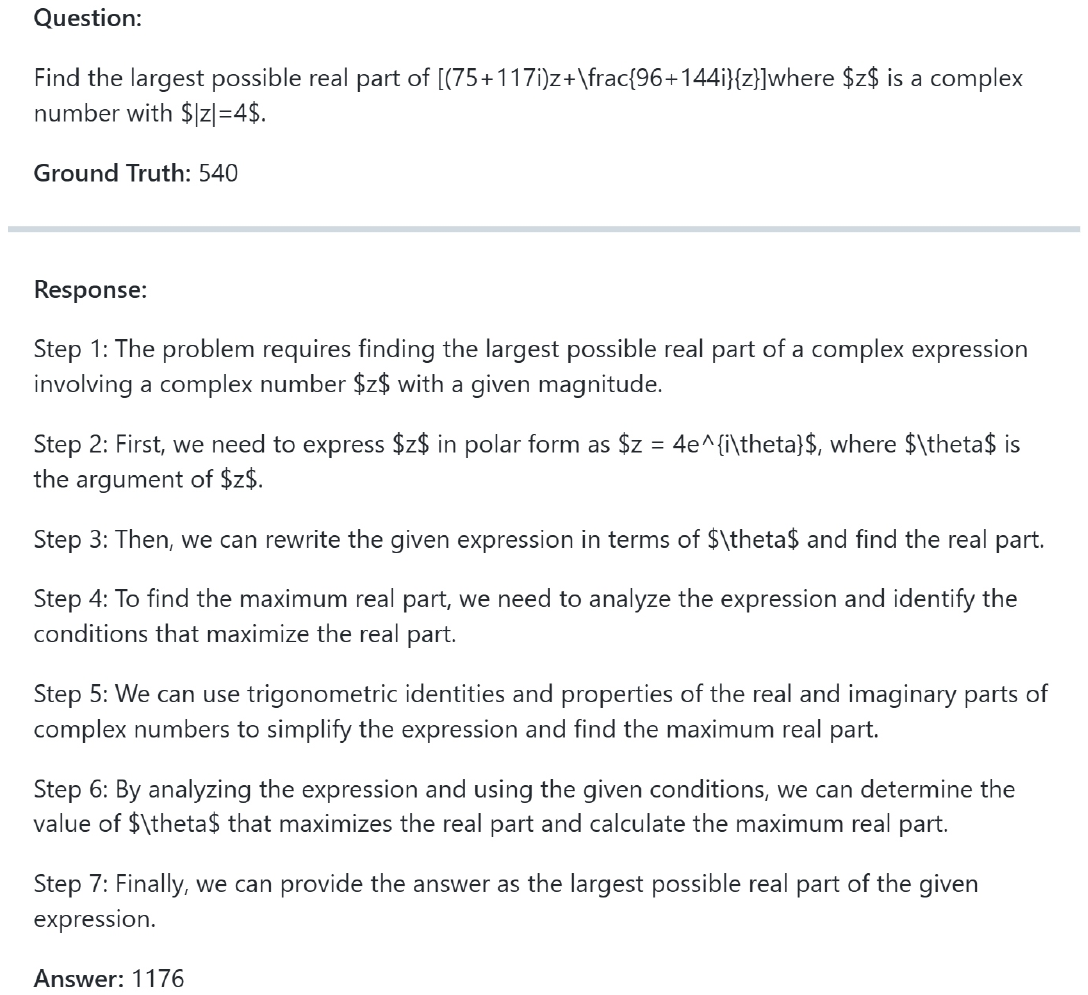}
    \caption{\textbf{Stage 1: Superficial Planning (Initial Step).} A baseline response from Qwen 2.5 32B. The model generates a "template-like" outline of the solution steps but fails to execute any actual mathematical derivation, leading to a hallucinated conclusion.}
    \label{fig:initial_output}
\end{figure}

\textbf{Stage 1: Superficial Planning (Initial Step).} As illustrated in \autoref{fig:initial_output}, the model at its initial step exhibits "superficial planning" behavior. It produces a template-like outline of the solution steps but fails to execute the actual mathematical derivation. This results in a brief response that lacks logical substance and often leads to an immediate hallucinated conclusion.

\begin{figure}[t] 
    \centering
    \includegraphics[width=\textwidth]{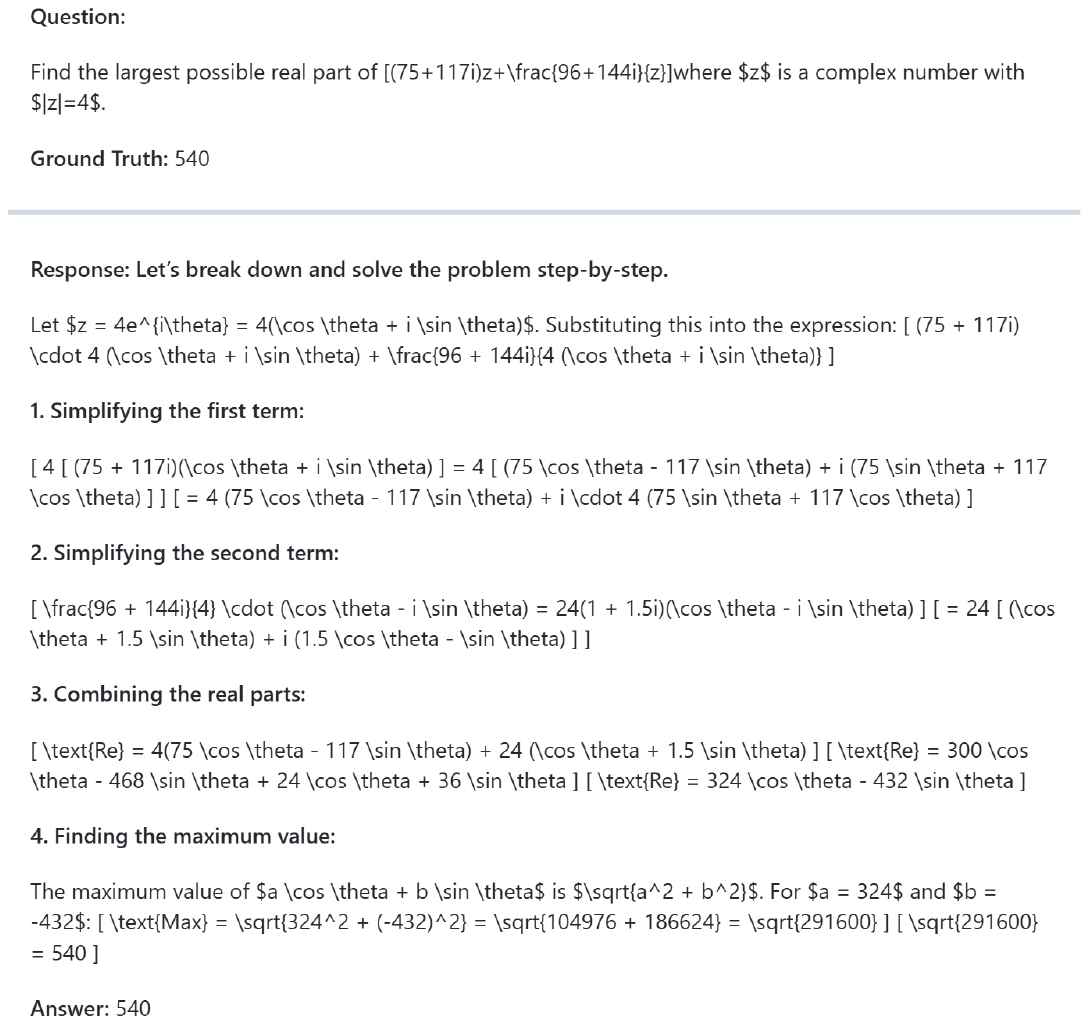}
    \caption{\textbf{Stage 2: Linear Execution (DAPO Convergence, and FIPO early stage).} While the model correctly executes a standard Chain-of-Thought (CoT) to reach the ground truth, its reasoning is \textit{strictly linear} and terminates immediately upon finding the first result, lacking any self-verification mechanism.}
    \label{fig:dapo_later}
\end{figure}

\textbf{Stage 2: Linear Execution (DAPO Convergence \& Early FIPO).} Across its entire training duration, DAPO remains situated within this stage (\autoref{fig:dapo_later}), evolving into a "linear executor" that can accurately follow a standard Chain-of-Thought (CoT) to reach the ground truth. However, its reasoning is inherently limited to a "single-pass" logic, where the generation terminates immediately upon finding the first result. This explains the length stagnation observed in \autoref{fig:response_length} for DAPO.

\begin{figure}[t] 
    \centering
    \includegraphics[width=\textwidth]{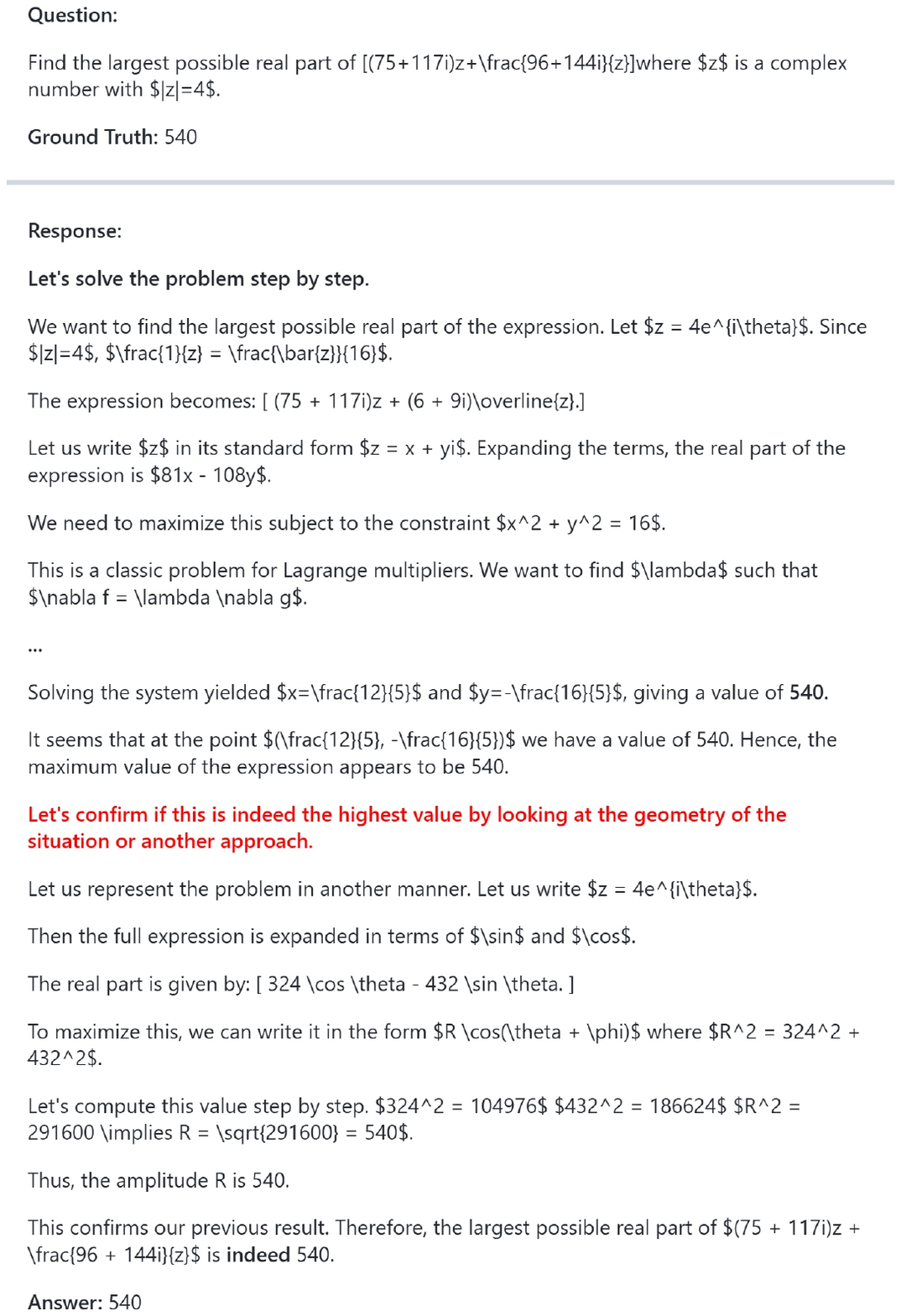}
    \caption{\textbf{Stage 3: Emergent Self-Reflection (FIPO Intermediate Stage).} A distinct behavioral shift occurs as FIPO training progresses. The model begins to utilize the expanded token budget for \textbf{spontaneous self-reflection}, cross-validating its initial conclusion through alternative mathematical paths.}
    \label{fig:checking_init}
\end{figure}

\textbf{Stage 3: Emergent Self-Reflection (Intermediate Stage FIPO).} As FIPO training progresses to the intermediate stage (\autoref{fig:checking_init}), a distinct behavioral shift occurs. The model begins to utilize the expanding token budget (growing in reponse length) for spontaneous self-reflection. After deriving an initial result, the model proactively initiates a verification phase, exploring alternative methodologies, such as switching from algebraic manipulation to geometric interpretation, to cross-validate its findings.

\begin{figure}[t] 
    \centering
    \includegraphics[width=0.95\textwidth]{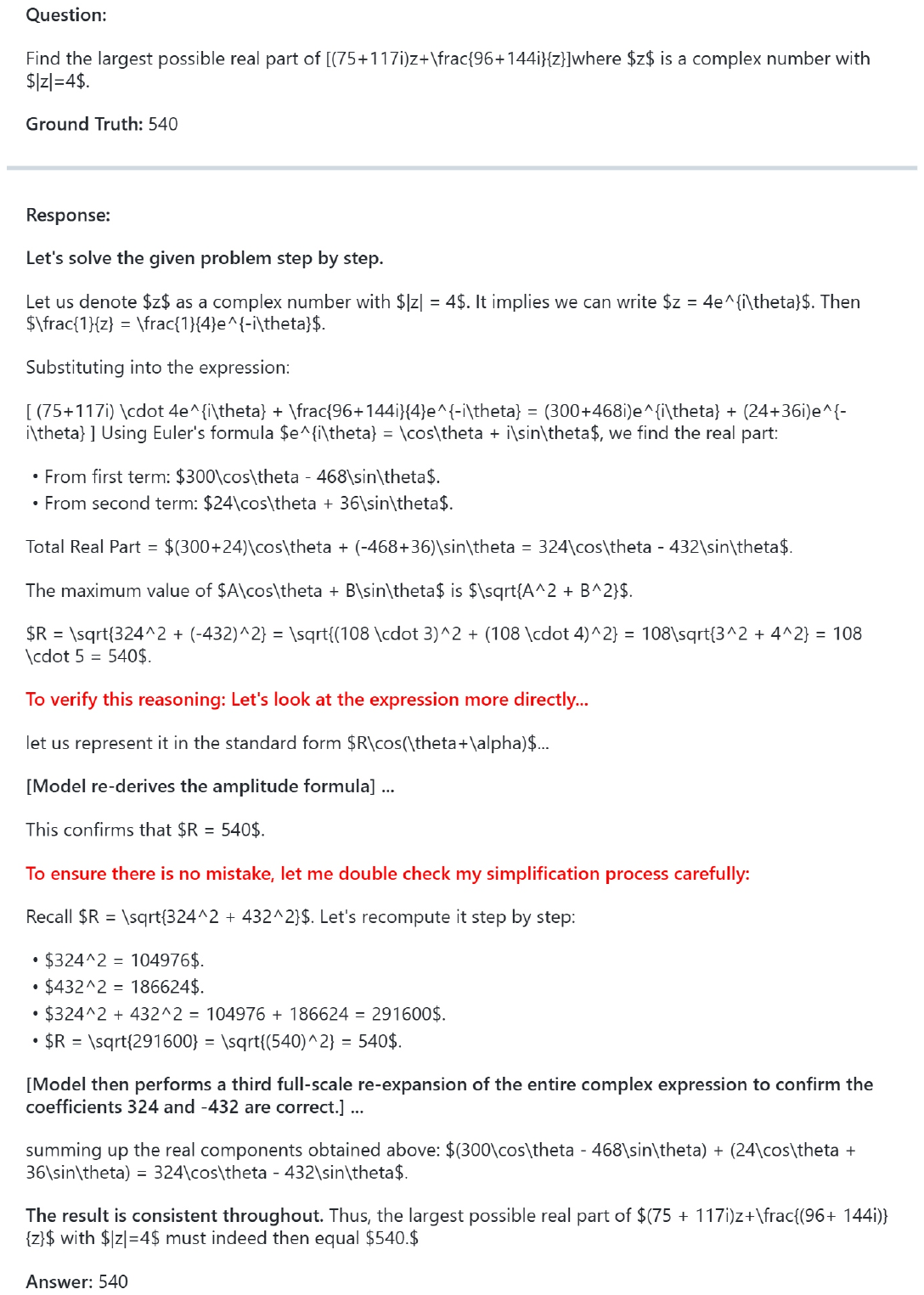}
    \caption{\textbf{Stage 4: Systematic Deep Reasoning (FIPO Late Stage).} In the late stages of training, the model converges on a \textbf{"compute-heavy" strategy}. It moves beyond simple reflection to perform rigorous \textbf{multi-pass auditing}—including symbolic re-derivation and granular arithmetic verification—to ensure better performance in complex reasoning tasks.}
    \label{fig:checking_deep}
\end{figure}

\textbf{Stage 4: Systematic Deep Reasoning (Late Stage FIPO).} In the later stages of training (\autoref{fig:checking_deep}), the model matures into a "compute-heavy" strategy that prioritizes analytical depth. The reasoning trajectory extends beyond simple reflection into a systematic audit. The model performs multiple passes of symbolic re-derivation and granular arithmetic verification (e.g., manually expanding large squares and square roots step-by-step). This spontaneous emergence of self-verification aligns with the inference-time scaling behaviors observed in advanced reasoning models, where the model treats length as a vital resource to ensure better performance.

\section{Discussions on failure trials and reproduction}
\label{app_sec: Reproduction_discussion}

\begin{figure}[ht] 
    \centering
    \includegraphics[width=\textwidth]{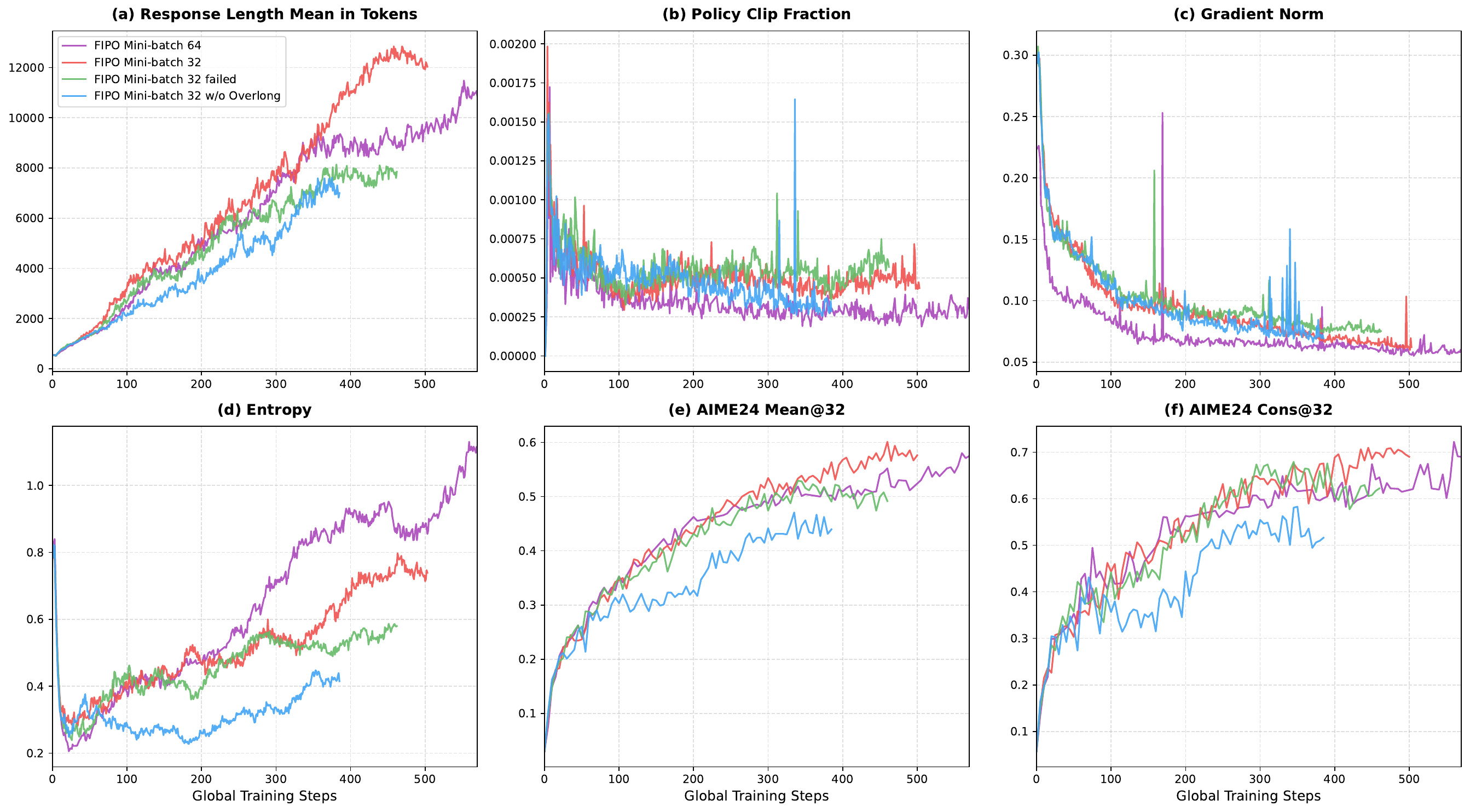}
    \caption{\textbf{Training Dynamics and Stability across Mini-batch Sizes on Qwen2.5-32B-Base.} 
    (a) \textbf{Response Length Mean:} While a mini-batch size of 32 can occasionally yield strong performance, it suffers from severe reproducibility issues and frequently encounters length stagnation. Notably, simply removing the overlong penalty (w/o Overlong) fails to resolve this bottleneck.
    (b) \textbf{Policy Clip Fraction:} In the successful mini-batch 32 trial, fewer tokens are clipped compared to the failed cases, a phenomenon that aligns with our findings in the \texttt{clip\_high} ablation study. Because a mini-batch size of 64 is closer to on-policy and suffers less from importance sampling (IS) weight fluctuations, it naturally results in fewer clipped tokens, thereby easing the scaling of response length and performance.
    (c) \textbf{Gradient Norm:} Similarly, the successful mini-batch 32 run maintains a noticeably stabler and slightly lower gradient norm than the failed ones, a critical stabilizing property inherently achieved by the mini-batch 64 baseline.
    (d) \textbf{Policy Entropy:} The successful case sustains higher entropy throughout the training, indicating healthier exploration capabilities.
    (e-f) \textbf{AIME24 Performance:} Benefiting from stable optimization and sustained exploration, the successful mini-batch 32 case reaches a peak Mean@32 of \textbf{approximately 60\%} and a \textbf{Cons@32 of 70\%}, which is highly comparable to the final performance of our mini-batch 64 baseline.}
    \label{fig:minibatch_stability}
    \vspace{-3mm}
\end{figure}

As noted earlier, we utilize a mini-batch size of 64 instead of 32 for our 32B model training and subsequent experiments due to its relative stability. To better understand this parameter choice, we detail below the specific training results and issues we encountered when employing a mini-batch size of 32. As illustrated in \autoref{fig:minibatch_stability}(a), while a mini-batch size of 32 can occasionally yield strong performance, it suffers from severe reproducibility issues. The most prominent failure mode is a severe deceleration in length growth, where the model struggles to effectively expand into longer reasoning trajectories. Notably, simply removing the overlong penalty fails to resolve this bottleneck, suggesting that the issue stems from deeper optimization dynamics rather than straightforward reward penalization.

We further interpret these findings through the underlying principles identified in our clip high ablation study (see \autoref{sec:cliphigh_ablation}). Although that ablation focuses on the clipping threshold rather than mini-batch size, it reveals a core optimization dynamic: when more tokens effectively participate in the policy update without being restricted by the clip boundary, the model gains significant momentum to expand its response length. Examining the policy clip fraction (\autoref{fig:minibatch_stability}(b)), we observe that the successful mini-batch 32 trial clips relatively fewer tokens than the failed ones. We deduce that when importance sampling (IS) weight fluctuations are reduced, a comparatively larger proportion of tokens remains unclipped and is fully included in the training signal. This valid inclusion is highly beneficial for overcoming the aforementioned length growth deceleration. Because a mini-batch size of 64 computes gradients over a broader sample set, it intrinsically stays closer to on-policy and suffers less from IS variance. As a result, it naturally avoids excessive token clipping, providing the sustained optimization momentum required to smoothly scale the reasoning length and performance.

The differences in optimization stability are also evident in the gradient norm and policy entropy. As shown in \autoref{fig:minibatch_stability}(c), the successful mini-batch 32 run maintains a noticeably stabler and slightly lower gradient norm than the failed runs. Furthermore, it sustains higher entropy throughout the training process (\autoref{fig:minibatch_stability}(d)), indicating healthier and more effective exploration capabilities without erratic divergence. Crucially, these stabilizing properties, controlled gradient updates and sustained, active exploration, are inherently achieved by the mini-batch 64 baseline, removing the need to rely on the stochastic success of smaller batch sizes. Ultimately, benefiting from stable optimization and consistent exploration, the successful mini-batch 32 case reaches a peak Mean@32 of approximately 60\% and a Cons@32 of 70\% (\autoref{fig:minibatch_stability}(e-f)). These metrics are highly comparable to the consistent performance of our mini-batch 64 baseline. Therefore, our adoption of the 64 mini-batch size is not driven by a higher theoretical performance ceiling, but rather by its capacity to reliably navigate the policy towards a high-quality reasoning manifold without falling into optimization traps.

It is crucial to distinguish this phenomenon from existing adaptive clipping techniques\citep{xi2025bapo,gao2025soft}. While adaptive clipping dynamically adjusts the threshold primarily to maintain trust-region stability and prevent policy collapse, our findings highlight a fundamentally different mechanism: the effective clip fraction acts as a structural valve for length expansion. By utilizing a mini-batch size of 64 to naturally stabilize IS weights, we allow a larger proportion of valid tokens to safely pass through a fixed boundary. Moreover, it is exactly under these stabilized optimization conditions that our proposed Future KL mechanism can fully exert its intended effect as the primary driver for reasoning training. By design, Future KL incorporates forward-looking signals to guide current token updates, thereby naturally encouraging the exploration of extended reasoning trajectories. However, this mechanism inherently relies on a continuous and intact flow of gradients across long sequences. The stabilized IS weights provides this exact foundation: \textbf{by preventing excessive token truncation, it ensures that the long-term exploratory signals from Future KL are successfully and smoothly propagated backward without being artificially interrupted}. Supported by this unhindered optimization momentum, Future KL explicitly pushes the policy to expand into deeper, high-quality reasoning manifolds, effectively preventing the model from collapsing back into premature, short-form responses.

\section{More details on Training Cost}

A naive implementation of Future KL requires computing a dense $(L, L)$ temporal decay matrix (where $L$ is the response length), resulting in an $\mathcal{O}(L^2)$ memory footprint that easily causes Out-Of-Memory (OOM) errors during long-trajectory reasoning training. To mitigate this, we implemented a chunk-based memory-efficient algorithm, as shown in \autoref{lst:future_kl}. By partitioning the response sequence into blocks of a fixed chunk size ($K$), we incrementally compute the distance masking and decay weights. The block-wise contributions are computed via parallel matrix multiplications of shape $(B, K) \times (K, L)$. This algorithm preserves the exact analytical outcome of the Future KL formulation while strictly bounding the peak memory complexity to $\mathcal{O}(B \cdot L + L \cdot K)$. Although the time complexity remains $\mathcal{O}(B \cdot L^2)$, the tensorized block operations are highly optimized on modern GPUs, effectively removing the memory bottleneck for scaling up reasoning lengths.

While standard GRPO computes the policy objective using strictly element-wise operations with an $\mathcal{O}(B \cdot L)$ time complexity, the integration of our Future KL mechanism naturally introduces an $\mathcal{O}(B \cdot L^2)$ temporal aggregation process. Consequently, this imposes a certain degree of computational overhead during the actor update phase. However, our chunked matrix multiplication implementation effectively vectorizes these operations, heavily leveraging the highly optimized dense MatMul capabilities of modern GPUs. Empirically, the wall-clock slowdown during the GRPO training iteration is relatively marginal and entirely acceptable. Most importantly, we argue that this modest trade-off in training speed is justifiable:\textbf{ it provides the dense, long-horizon credit assignment necessary to scale up complex reasoning}, a fundamental bottleneck that standard $\mathcal{O}(B \cdot L)$ GRPO inherently struggles to overcome. Naturally, while our chunked implementation effectively addresses the memory constraints, more sophisticated computational optimizations remain possible.

\begin{figure}[htbp]
\begin{lstlisting}[caption={Python code for computing memory-efficient chunked Future KL}, label={lst:future_kl}]
import torch

def compute_future_kl_chunked(D, M, K, tau):
    # Calculate the decay factor
    gamma = 2 ** (-1 / tau)
    
    # Mask out invalid tokens and filtered anomalies
    D = D * M
    
    # Initialize the Future KL accumulator
    F = torch.zeros_like(D)
    L = D.shape[1]
    
    # Column vector of query positions (L x 1)
    i = torch.arange(L, device=D.device).unsqueeze(1)
    
    for j_start in range(0, L, K):
        j_end = min(j_start + K, L)
        
        # Row vector of chunk positions (1 x K_cur)
        j = torch.arange(j_start, j_end, device=D.device).unsqueeze(0)
        
        # Broadcasted distance matrix (L x K_cur)
        Delta = j - i
        
        # Decay weight block
        W = (gamma ** torch.clamp(Delta, min=0)) * (Delta >= 0).float()
        
        # Extract KL values for the current chunk
        V = D[:, j_start:j_end]
        
        # Parallel matrix multiplication update
        F += torch.matmul(V, W.T)
        
    return F
\end{lstlisting}
\end{figure}

\end{document}